%% file: main.tex
\documentclass[runningheads]{llncs}

\usepackage{eccv}

\usepackage{eccvabbrv}

\usepackage[utf8]{inputenc} % allow utf-8 input
\usepackage[T1]{fontenc}    % use 8-bit T1 fonts
\usepackage{url}            % simple URL typesetting
\usepackage{booktabs}       % professional-quality tables
\usepackage{amsfonts}       % blackboard math symbols
\usepackage{nicefrac}       % compact symbols for 1/2, etc.
\usepackage{microtype}      % microtypography
\usepackage{xcolor}         % colors
\usepackage{graphicx}
\usepackage{algpseudocode}
\usepackage{amssymb}
\usepackage{caption}
\usepackage{amsmath}
\usepackage{subcaption}
\usepackage{pifont}
\usepackage[capbesideposition=outside,capbesidesep=quad]{floatrow}
\newfloatcommand{capbtabbox}{table}[][.45\textwidth]
\newcommand{\xmark}{\text{\ding{55}}}

\usepackage{graphicx}
\usepackage{booktabs}

\usepackage[accsupp]{axessibility}  % Improves PDF readability for those with disabilities.

\usepackage{hyperref}

\usepackage{orcidlink}

\begin{document}

% ---------------------------------------------------------------
% TODO REVIEW: Replace with your title
\title{
Made to Order: Discovering monotonic temporal changes via self-supervised video ordering 
}

% TODO REVIEW: If the paper title is too long for the running head, you can set
% an abbreviated paper title here. If not, comment out.
\titlerunning{Discovering monotonic temporal changes via self-supervised video ordering}

% TODO FINAL: Replace with your author list. 
% Include the authors' OCRID for the camera-ready version, if at all possible.
\author{Charig Yang\inst{1}\orcidlink{0009-0003-7044-1901} \and
Weidi Xie\inst{1,2}\orcidlink{0009-0002-8609-6826} \and
Andrew Zisserman\inst{1}\orcidlink{0000-0002-8945-8573}}

% TODO FINAL: Replace with an abbreviated list of authors.
%\author{
%Charig Yang\inst{1} \and
%Weidi Xie\inst{1,2} \and Andrew Zisserman\inst{1}}

% TODO FINAL: Replace with an abbreviated list of authors.
\authorrunning{C.~Yang et al.}
% First names are abbreviated in the running head.
% If there are more than two authors, 'et al.' is used.

% TODO FINAL: Replace with your institution list.
\institute{Visual Geometry Group, University of Oxford\and
CMIC, Shanghai Jiao Tong University \\
\email{\{charig,weidi,az\}@robots.ox.ac.uk} \\
\href{https://charigyang.github.io/order/}{\texttt{https://charigyang.github.io/order/}}}

\maketitle

\begin{abstract}
Our objective is to discover and localize monotonic temporal changes in a sequence of images. To achieve this, we exploit a simple proxy task of ordering a shuffled image sequence, with `time' serving as a supervisory signal, since only changes that are monotonic with time can give rise to the correct ordering. We also introduce a transformer-based model for ordering of image sequences of arbitrary length with built-in attribution maps. After training, the model successfully discovers and localizes monotonic changes while ignoring cyclic and stochastic ones. We demonstrate applications of the model in multiple domains covering different scene and object types, discovering both object-level and environmental changes in unseen sequences. We also demonstrate that the attention-based attribution maps function as effective prompts for segmenting the changing regions, and that the learned representations can be used for downstream applications. Finally, we show that the model achieves the state-of-the-art on standard benchmarks for image ordering. %ordering a set of images.
\end{abstract}
%\maketitle

\begin{center}
    \centering
  \includegraphics[width=\textwidth]{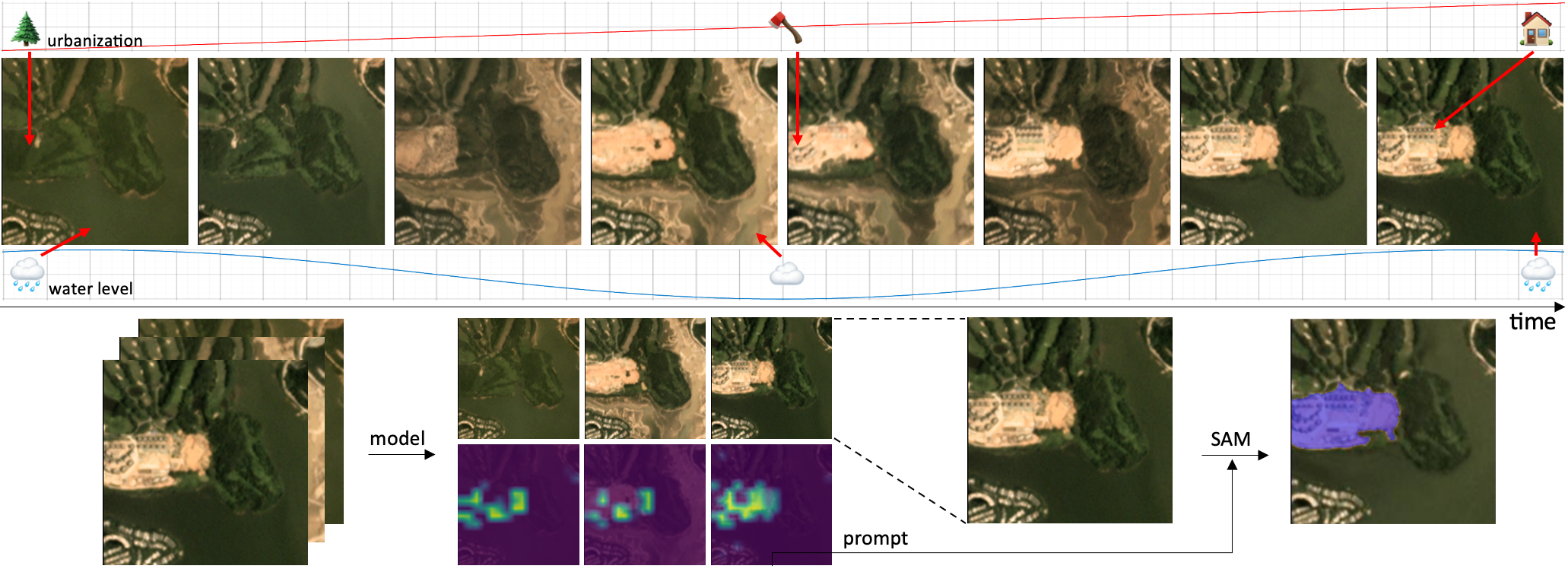}

  \captionof{figure}{ {\textbf{Localizing monotonic temporal changes.}
  Top: satellite images (ordered left to right) taken months apart, containing several changes -- some are monotonic (\eg urbanization), while others are seasonal/cyclic (\eg water level).
  Bottom: Our model's attribution map prediction on the sequence is able to localize the regions with monotonic temporal changes (in green), while being invariant to the seasonal and sporadic changes. 
  The model is trained with no manual supervision, generalises to unseen sequences (as here), and the attribution map can also be used as a prompt to obtain segmentation.}}
  \label{fig:teaser2}
\end{center}%

\input{sec/1_intro}
\input{sec/2_related_work}

\input{sec/3_method}

\input{sec/4_experiments}

\input{sec/5_results}

\input{sec/6_conclusion}

\bibliographystyle{splncs04}
\bibliography{main}

\clearpage
\include{sec/X_suppl}

% WARNING: do not forget to delete the supplementary pages from your submission 

\end{document}

%% file: sec/1_intro.tex
\section{Introduction}

In the image sequence in Figure \ref{fig:teaser2}, there exists numerous changes over time, though many of these are seasonal, and hence distracting for long-term monitoring applications. As humans, 
we not only observe what is changing, but also reason about which changes are correlated monotonically with time and which ones are not. In this paper, we introduce a new task of automatically identifying temporally correlated changes in an image sequence, while being invariant to other changes. More specifically we wish to {\em discover} and {\em localize} the image regions where the change is correlated with time. Our motivation is to go beyond just detecting changes, but also discovering what changes are relevant over a period of time, and to explore the potential applications that this task could enable.

To achieve this, we propose a self-supervised proxy task, with time serving as a supervisory signal: the task is simply to order a shuffled sequence of video frames. The insight is that {\textbf{frames are only orderable if changes are monotonic}, therefore} if a model can order the video frames, it would have learned to identify relevant (monotonic temporal-correlated) cues while disregarding other changes. The trained model can then be employed for video analysis applications where the goal is to identify \textbf{changes} over time, 
such as developments or deforestation in  satellite imagery (whilst ignoring seasonal variations) (see Figure \ref{fig:teaser2}) or aging signs in medical images. It can also be employed for detecting and tracking monotonic object \textbf{motion} over time, such as shadows caused by the movement of the sun or animals moving smoothly across the scene.

It is worth noting that temporal ordering has previously been used as a proxy task for self-supervised representation learning, with the learnt representations then finetuned for downstream tasks, such as video action recognition~\cite{wei2018learning,misra2016shuffle, lee2017unsupervised, fernando2017self} 
(though due to the disparity between the proxy and downstream tasks, the effectiveness of learnt visual representation from temporal ordering has been unsatisfactory compared to other proxy tasks~\cite{chen2020simple, han2020coclr, grill2020bootstrap, he2021mae}).
In contrast, the objective in this paper is to use ordering as a proxy task to directly train a model for discovering and localizing monotonic changes in video sequences, without any subsequent supervised finetuning. In this sense, we are similar in spirit to previous works that use self-supervision to directly solve tasks, such as \cite{jabri2020spacetime, lai2020mast, bian2022learning,yang2021self} that targets tracking and segmentation by training on proxy tasks.

In order to harness the ordering proxy task, 
we introduce a transformer-based model that is able
(i) to perform  {\em ordering} on natural images sequences, 
and, more importantly, (ii) to provide {\em attribution} by localizing the evidence that gives rise to its prediction. 
Specifically, we use a DETR-like transformer encoder-decoder architecture where the queries in the decoder are cast as an {\em ordering index}. 
The architecture is designed to allow an attribution map to be obtained directly as part of the proxy training. Once the model has been trained for a particular setting, such as change detection in satellite image sequences, 
then it can generalize to unseen videos in the same setting, 
requiring only a forward pass to predict the localization of the monotonic temporal changes from the attribution map.

To summarize, in this paper, we make the following four contributions:
(i) we introduce a new task of discovering and localizing monotonic temporal changes in image sequences, and {establish the link that temporal ordering can be used as a self-supervised proxy task for training; (ii) we introduce a flexible transformer-based architecture for ordering that can also automatically localize monotonic changes;}
(iii) once trained on a setting (such as satellite images), 
the model is able to discover the changes correlating monotonically with time in unseen image sequences in the same setting, and we demonstrate several situations where this can be applied. 
Finally, (iv) we demonstrate that the trained model is able to order novel image sequences surpassing the performance of previous approaches for ordering on standard benchmarks.

%% file: sec/2_related_work.tex
\section{Related Work}

\noindent\textbf{Video self-supervised learning} has become increasingly popular in computer vision. Most research in this area focuses on representation learning, with an emphasis on downstream performance after supervised fine-tuning or linear probing. In contrast, there is a less explored direction that goes beyond representation learning to directly learning a useful task under the self-supervised learning setting, such as depth \cite{zhou2017unsupervised, godard2019digging}, optical flow \cite{meister2018unflow, liu2019selflow}, correspondence \cite{wang2019learning} and sound localization~\cite{afouras2020self, arandjelovic2018objects,Chen21,liu2022exploiting}. Our work in this paper builds upon such paradigm, that enables to deploy the model to downstream tasks without the need for labels.

\noindent\textbf{Self-supervised learning from time.} 
This work involves using temporal ordering as a supervisory signal. Alternative sources of supervision is to leverage other cues, such as speed~\cite{benaim2020speednet}, uniformity~\cite{yang2022s}, and the arrow of time~\cite{wei2018learning}. The closest kin to our work involves using temporal sequencing as supervision~\cite{misra2016shuffle, fernando2017self, lee2017unsupervised, fernando2015modeling}, though their primary focus is on representation learning. In this work, we focus on advocating temporal ordering as a useful task on its own, showing that localization can emerge by using time as supervisory signal. We highlight the differences between our work and others in the Supplementary Material.

\noindent\textbf{Ordering} has been a longstanding task in computer science, 
dating back to sorting algorithms. 
In machine learning, it is a relevant task in both language~\cite{chen2016neural, cui2018deep} and vision~\cite{misra2016shuffle, basha2012photo, zhukov2020learning, shvetsova2023learning}.
For images, ordering has also been treated as a pretext for self-supervised pretraining, such as jigsaw puzzles~\cite{noroozi2016unsupervised},
or as a task of interest, for example, image sequencing~\cite{(Basha)_2013_ICCV, basha2012photo, zarrabi2018crowdcam, sevilla2021only, kim2023learning}. 
There is also some interest in the literature that focuses on differentiable sorting algorithms \cite{petersen2021differentiable, petersen2022monotonic, cuturi2019differentiable, grover2019stochastic, Brown20}, though they mostly focus on algorithmic developments, such as differentiable loss function and black-box combinatorial optimisation. 
While in this paper, we make contributions towards the architecture by introducing a transformer-based ordering model, which allows ordering of arbitrary-length image sequences, with built-in visualisation via attribution maps.

\noindent\textbf{Attribution localization}, 
specifically in the case where explicit supervision is not given, has been of interest in the vision community. 
In ConvNets, attribution methods attempt to look into the network to find out where it is seeing \cite{zhou2016learning,fong2017interpretable}. 
In transformers, several methods have been proposed to look into the attention on the \texttt{[CLS]} tokens \cite{abnar2020quantifying}. Instead of these implicit localization, several methods have also carefully designed the architecture so that localization emerges explicitly despite not being trained on, such as in sound localization \cite{afouras2020self, arandjelovic2018objects,Chen21,liu2022exploiting}.
This work follows the latter paradigm, while using the self-supervision from ordering.

\noindent\textbf{Change detection} has also been studied in computer vision. Many works look at changes in the image domain \cite{Sachdeva_WACV_2023, Sakurada2015ChangeDF}, and across different applications from construction monitoring \cite{Stent2015DetectingCF}, satellite imaging \cite{mall2023change}, to medical imaging \cite{patriarche2004review}.
Other works associate short-term changes with motion, 
and use motion as a cue to discover moving objects \cite{Bideau16,Bideau16a,lamdouar2020betrayed,yang2021self,Lamdouar21,xie2022segmenting}. We differ from these lines of work in that we are mostly concerned with temporally coherent changes at different timescales, 
which may go beyond object level and not associated with motion.

%% file: sec/3_method.tex
\begin{figure}[t]
  \centering
  \includegraphics[width=\textwidth]{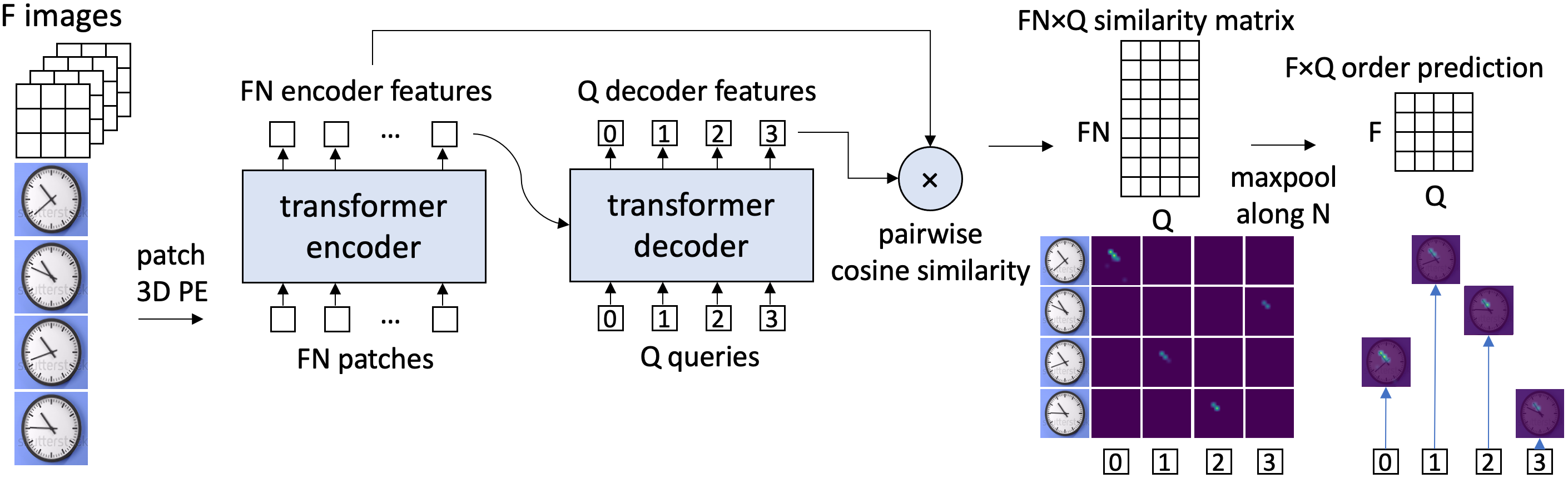}
  \caption{\textbf{Network architecture.} For an unordered sequence of $F$ frames each with $N$ patches, the transformer encoder takes in all $FN$ patches as input, and outputs $FN$ features. 
  The transformer decoder takes in $Q$ learnable queries, each corresponding to an ordinal position, 
  and the encoder output for cross-attention, 
  resulting in $Q$ features for output. 
  A $FN \times Q$ cosine similarity matrix is constructed between all pairs of features from the encoder and decoder outputs, and the spatial max-pooling over this matrix reveals the $F \times Q$ order predictions. The ordering can simply then be obtained by taking an argmax along each query axis. 
  In the example sequence, the hour hand is correlated monotonically with time, and appears in the attribution map.}
%\az{Clock change is too perhaps too subtle. Comment on clock example in caption -- that the hour hand is correlated monotonically with time, and appears in the similarity matrix (or map)}  }
  \label{fig:arch}
  
\end{figure}
\section{Method}

\subsection{Problem Formulation}

Our goal is to train a vision model to localize the changes in an image sequence that correlate monotonically with time. 
As a subsidiary goal, the model should also be able to order the image sequence.

Formally, given set of images, the model should output an {\em attribution map} $\mathbf{S}_{\text {att}} \in \mathbb{R}^{F\times H\times W}$ and an ordering $y_{\text{order}} \in \mathbb{Z}^{F} : y_{\text{order},i} \in \{0,1,2,...,F-1\}$ as:
$$
y_{\text{order}},  \mathbf{S}_{\text {att}} = \Phi(\mathcal{I}_0, \mathcal{I}_1, ..., \mathcal{I}_{F-1}) 
$$
where $\mathcal{I} \in \mathbb{R}^{C\times H\times W}$ represents the input images. We show that we can train the model  $\Phi$  via self-supervised learning on a proxy task, namely, ordering an arbitrary sequence of $F$ images shuffled from a temporal sequence. 

\subsection{Ordering architecture}

To address this problem, 
we propose a simple yet novel transformer-based architecture, 
as shown in Figure~\ref{fig:arch}. 
The architecture comprises a transformer encoder~($\Phi_{\text{enc}}$) that encodes the image patches, and a transformer decoder~($\Phi_{\text{dec}}$) that encodes the ordering. 
To obtain an attribution map, we simply compute the pairwise cosine similarity between features from the encoder and queries from the decoder. We can then perform a max pooling operation across patches of the same image to get the ordering prediction.

\noindent {\bf Transformer Encoder~($\Phi_{\text{enc}}$).}  
To process an unordered sequence of $F$ images, 
{\em i.e.}, $\mathcal{X} = \{\mathcal{I}_0, \mathcal{I}_1, ..., \mathcal{I}_{F-1}\}$,
we start by dividing each frame $\mathcal{I} \in \mathbb{R}^{ C\times H\times W}$ into 2D patches of size $(P, P)$, resulting in $N=HW/P^2$ patches per frame and $FN$ patches in total. 
Following the vision transformer approach, we flatten each patch using a learnable projection layer to $D$ dimensions and add 3D positional encoding (spatial and frame) to each patch. 

It is important to note that the frame positional encoding does not contain absolute temporal information since the frames are unordered, but it allows the patches to identify whether they belong to the same frame. 
As a result, after patchifying the input sequence, 
it ends up with a tensor of $\mathbf{x} \in \mathbb{R}^{F\times N \times D}$, 
which is then fed into a transformer encoder. 
The key difference to the standard vision transformer 
is that we output all the features,
{\em i.e.}, $\mathbf{x}_{\text{enc}} \in \mathbb{R}^{F\times N \times D}$ 
instead of using a \texttt{[CLS]} token. 
In summary, we can express this procedure as $\mathbf{x}_{\text{enc}} = \Phi_{\text{enc}}(\mathcal{I}_0, \mathcal{I}_1, ..., \mathcal{I}_{F-1})$.

\noindent {\bf Transformer Decoder~($\Phi_{\text{dec}}$).} 
The transformer decoder is composed of $Q$ learnable queries $\mathbf{q} \in \mathbb{R}^{Q\times D}$, with each corresponding to an ordering position $(0, 1, ..., Q-1)$. 
The task for the transformer decoder is to align the query vector with the encoder feature that demonstrates the correct temporal order. These queries iteratively attend the visual outputs from the encoder with cross-attention in the standard transformer decoder. We denote the output of the decoder as, $\mathbf{x}_{\text{dec}} \in \mathbb{R}^{Q \times D} = \Phi_{\text{dec}}(\mathbf{q}, \mathbf{x}_{\text{enc}})$. In practice, $Q=F$.

\noindent {\bf Cosine similarity matrix~(S).}
Recall that we now possess two sets of features: encoder features $\mathbf{x}_{\text{enc}} \in \mathbb{R}^{F\times N \times D}$ and decoder features $\mathbf{x}_{\text{dec}} \in \mathbb{R}^{Q \times D}$. We then compute the pairwise cosine similarity matrix $\mathbf{S} \in \mathbb{R}^{F\times N\times Q} : [\mathbf{S}]_{i,j} = \cos(\mathbf{x}_{\text{enc},i}, \mathbf{x}_{\text{dec},j}) \in [-1, 1]$ between each $i$ of the $F\times N$ features in $\mathbf{x}_\text{enc}$ and each $j$ of the $Q$ features in $\mathbf{x}_\text{dec}$, where $\cos(\cdot,\cdot)$ denotes the cosine similarity function.

Given the similarity matrix, we want to obtain (i) the ordering of the frames and (ii) the attribution map that indicates the spatial evidence within each frame that gives rise to ordering. The matrix $\mathbf{S} \in \mathbb{R}^{F\times N\times Q}$ consists of $F\times Q$ different spatial maps of size $N$, each indicating the attention between each pair of queries~($j \in Q$) and images~($i \in F$). %\weidi{the last sentence is unclear.}

\noindent {\bf Order prediction.} To obtain the order predictions, we perform spatial max-pooling over the patches of each frame (along the N dimension), to obtain $\hat{y} \in \mathbb{R}^{F\times Q} = \max_{i\in N} \mathbf{S}_i$. This max-pooling is designed to create an information bottleneck -- the query has to attend to the correct token(s) within the correct image in order to predict the order correctly. The resulting matrix serves as a predictor for the position of each query in the ordering. We then apply a softmax along the query axis of the matrix to get the probability scores for each query.

\noindent {\bf Attribution map.} Among the $F\times Q$ different spatial maps, we are only interested in the ones that correspond to the correct ordering. For each query $j \in Q$, we select one map $i \in F$ that has the maximum activation, {\em i.e.} $ i = \arg \max \hat y_{j}$ resulting in $Q$ maps. We then rearrange and resize each map of $N$ patches back to the original resolution, resulting in $\mathbf{S}_{\text {att}} \in \mathbb{R}^{Q\times H\times W}$. Notably, this localization can be achieved without the need for additional fine-tuning, supervision, or post-hoc attribution methods~\cite{fong2017interpretable, abnar2020quantifying}.

\subsection{Training and inference}

\noindent {\bf Temporal loss. } Given the ground-truth order $y \in \mathbb{Z}^{Q} : y_{i} \in \{0, 1, ..., F-1\}$, the model can be trained via binary cross-entropy loss. {Specifically, we convert the ground-truth order into a binary permutation matrix.} %matrix where each column is a one-hot vector indicating its position,
%e.g. $(1,0,2)$ as $\big(\begin{smallmatrix}
%  0 & 1 & 0\\
%  1 & 0 & 0\\
%  0 & 0 & 1
%\end{smallmatrix}\big)$. 
With some notation abuse, we still call this matrix $y \in \mathbb{Z}^{F\times Q}$. The forward loss is then simply the elementwise binary cross-entropy between the two matrices: $\mathcal{L}_f(y, \hat{y}) = \frac{1}{FQ}\sum_{i\in F}\sum_{j\in Q} \text{cross-entropy}(\hat y_{ij}, y_{ij})$. 

In practice, we find that allowing reversibility in the loss aids with training, as many changes
are reversible in nature without prior knowledge of the arrow of time~\cite{wei2018learning} (the sequence could equally be ordered from first to last, or last to first). To allow this, we calculate the loss as the minimum of the loss for both forward and backward sequences, {\em i.e.}\ $\mathcal{L}_r = \min(\mathcal{L}_f(y,\hat{y}), \mathcal{L}_f(\text{reverse}(y),\hat{y}))$. 
{This loss is zero when the model predicts the order correctly in either direction.}

\noindent {\bf Inference. } 
At inference time, we simply take the argmax along each query axis as the order prediction, 
that is, $y_{\text{order}} \in \mathbb{Z}^{Q} : y_{\text{order},j}  = \arg \max_{i \in F} \hat{y}_{i, j} $
In other words, each query picks the image that contains the maximum activation for its query, as illustrated in the bottom-right corner of Figure \ref{fig:arch}.

\subsection{Discussion}

\label{sec:diss}

\noindent {\bf Generalization to different sequence lengths. }
Our architecture is designed to handle sequences of arbitrary, possibly unequal length during training and inference, without the need to re-design or train separate models for each sequence length. At training time, we assume there is a maximum number of images, thus initialize a total of $F_{\text{max}}$ learnable queries in the transformer, {\em i.e.}, $\mathbf{q} \in \mathbb{R}^{F_{\text{max}}\times D}$. While the model handles a sequence of $F$ images, with $F\leq F_{max}$, it only uses the first $F$ queries as input to the decoder, ignoring the rest. This approach enables each query to represent its positioning $(0,1,...,F-1)$, making it generalizable to different lengths during both training and testing. However, the model will not generalize to lengths above $F_{\text{max}}$.

\noindent {\bf Avoiding trivial solutions. }
There are two factors that we need to account for: camera motion and video compression artifacts. Camera motion can be smooth or uniform over a short time gap, which can result in an uninteresting cue. 
To address this, we apply a small random cropping on each frame in settings where the time gap between frames is small ({\em i.e.} $<1$s).
This slight jittering helps to prevent the model from learning trivial solutions. 
We note that this does not degrade the performance even if camera motion is absent. 
Another factor that can give rise to trivial solutions is inter-frame video compression artifacts. To address this, we follow conventional wisdom~\cite{iashin2022sparse, wei2018learning} and use H.264 formatting for all videos, 
thus minimizing compression artifacts and preventing trivial solutions.

\noindent {\bf From localization to segmentation.} While the attribution map is useful, some applications may benefit from going beyond just localization. Here, we propose {three} solutions. 
(i) We can directly obtain segmentation at patch-level granularity by thresholding the attribution map, together with minimal post-processing namely averaging across frames and removing small contours. 
{
(ii) We replace the linear projection layer on image patches with a pretrained DINOv2 to enhance the feature quality.
}
%We note that this is crude as it does not distinguish pixels within a patch.
(iii) Alternatively, we can use the highest activation points ({\em i.e.} centre pixels of patches) as prompt for the pretrined Segment Anything model (SAM)~\cite{kirillov2023segany}, to obtain pixel-level segmentation masks. 
%We use the pretrained SAM model without fine-tuning. 

%% file: sec/4_experiments.tex
\begin{figure*}[t]
  \centering
  \includegraphics[width=\textwidth]{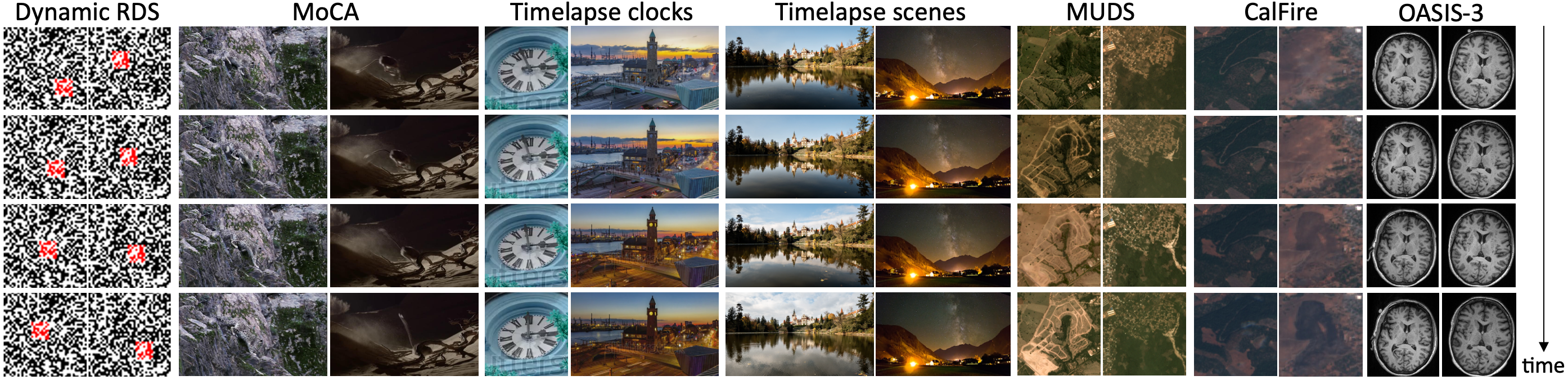}
  \caption{\textbf{Sequence datasets.} From left to right: dynamic Random Dot Stereograms (RDS) (moving dots colored only for illustration), moving camouflaged animals (MoCA), timelapse clocks (cropped/full), timelapse scenes, MUDS, CalFire, OASIS-3.
  }
  \label{fig:dset}
\end{figure*}

\section{Experiments}
\subsection{Video datasets}
To leverage the inherent temporal information in videos,
we sample a sequence from a video, shuffle the frames, 
then train the model with the original ordering as groundtruth. 
This enables the attribution map to identify monotonic changes that contribute to the ordering while disregarding other changes. 
Our study explores sequences across multiple domains, each with distinct cues of interest, which we summarize in Table~\ref{tab:results} and illustrate sample sequences in Figure~\ref{fig:dset}. 

\noindent\textbf{Dynamic random dot stereograms} are a type of image sequence that features a box of random dots moving smoothly over a background of random dots, originally used as a pair to demonstrate stereoscopic motion~\cite{neff1985electronics}. While the individual images may appear random, the box is visible when viewed in sequence. 
We generate a synthetic dataset of controllable dynamic random dot stereograms~(Dynamic RDS) to test the model's ability to detect subtle relative cues. Since this a synthetic dataset, we know the ground-truth of the box's motion, so can compare this with the predictions.

\noindent\textbf{Moving camouflaged animals (MoCA)} ~\cite{lamdouar2020betrayed}
was constructed from videos of camouflaged animals.
We use this dataset to investigate the use of short-term object locomotion as a cue, particularly where it is challenging to distinguish the object from the background. To accomplish this, we follow~\cite{yang2021self} and focus on the subset of 88 videos in which the animals are in motion. To evaluate on localisation, we assume that the change is object-level due to animal motion, and use the annotated object bounding box as the ground-truth for localization.

\noindent\textbf{Timelapse analog clocks}.
Our study examines real-world scenes that feature both absolute cues (time on clocks) and relative cues (scene changes). 
To accomplish this, we utilize the Timelapse dataset~\cite{yang2022s} in two ways. Firstly, we use the entire dataset of 2,511 videos that features cropped clocks. Secondly, we create a subset consisting of 260 outdoor scenes with static cameras where the clock occupies only a small area of the scene.

\noindent\textbf{Timelapse scenes.} {We gathered outdoor timelapse videos from WebVid-10M dataset \cite{Bain21} with `timelapse' as the search query}. The dataset comprises 180 static videos, and our aim is to investigate the cues that the model can extract from the scene to learn its order, 
as there are no specific absolute cues present.

\subsection{Temporal image sequences}
We also show effectiveness of our approach for image sequences that are collected over a period of time, including satellite images and longitudinal medical data.

\noindent\textbf{Multi-temporal urban development dataset (MUDS)} \cite{van2021multi} is a dataset that contains 80 aerial satellite image sequences captured monthly over a two-year period.
We aim to identify geographical changes that occur over time, including deforestation and urbanization, while ignoring other changes.
As there are no existing datasets and benchmarks for this task, we hand-label 60 sub-sequences on the test set for segmentation masks where changes are monotonic, and call this evaluation set \textbf{Monotonic MUDS}. Samples are shown in Figure~\ref{fig:mmuds}.
We use this dataset to evaluate localization and segmentation performance of the model trained on the MUDS dataset.

\noindent\textbf{CalFire} \cite{mall2022change} is a satellite dataset tracking wildfires in California. It also contains other events, such as snowfall, new construction, changes in water level, that we aim to discover. We pre-process by removing scenes with significant cloud cover.

\noindent\textbf{OASIS-3} \cite{lamontagne2019oasis} contains longitudinal MRI scans taking 1-4 years apart to study how brains age. We select sequences with 3 or more scans, resulting in 134 sequences. Following \cite{kim2023learning}, we perform affine registration and use the centre slice.

\begin{table}[t]
\footnotesize
\centering
\setlength\tabcolsep{3pt}
\begin{tabular}{lccc|cc}
\toprule dataset & $\Delta t$ & cue & seq (trn/test)  & EM & EW  \\ \midrule
Dynamic RDS   & <1s & motion  & $\infty$ &  99.8 & 99.9 \\ % & --- & --- \\ 
MoCA   & <1s & motion & 75/13 &  82.0 & 90.6\\
Clocks (cropped)  & $\sim$1m & clock & 2011/500  & 62.5 & 73.0 \\
Clocks (full)   & $\sim$1h & clock/scene  &210/20 & 55.0 & 74.3 \\
Timelapse scenes   & $\sim$1h & scene  &130/50  &61.8 & 79.4\\
MUDS  & 1mo & landscape  &60/20 & 56.4 &  69.6 \\
CalFire  & 1mo & landscape &800/276& 76.6 &  87.5  \\
OASIS-3   & 1y & brain  &  100/34&84.3& 89.1  \\
 \bottomrule
\end{tabular}
\caption{\textbf{Dataset attributes and ordering results.} This table shows different datasets and their attributes, as well as the ordering results on the test (unseen) sequences on exact match (EM) and elementwise (EW) metrics.}
\label{tab:results}

\end{table}

\subsection{Image ordering datasets}
\label{sec:image}
\begin{figure}[t]
\centering
\begin{subfigure}{.5\textwidth}
  \centering
  \includegraphics[width=\linewidth]{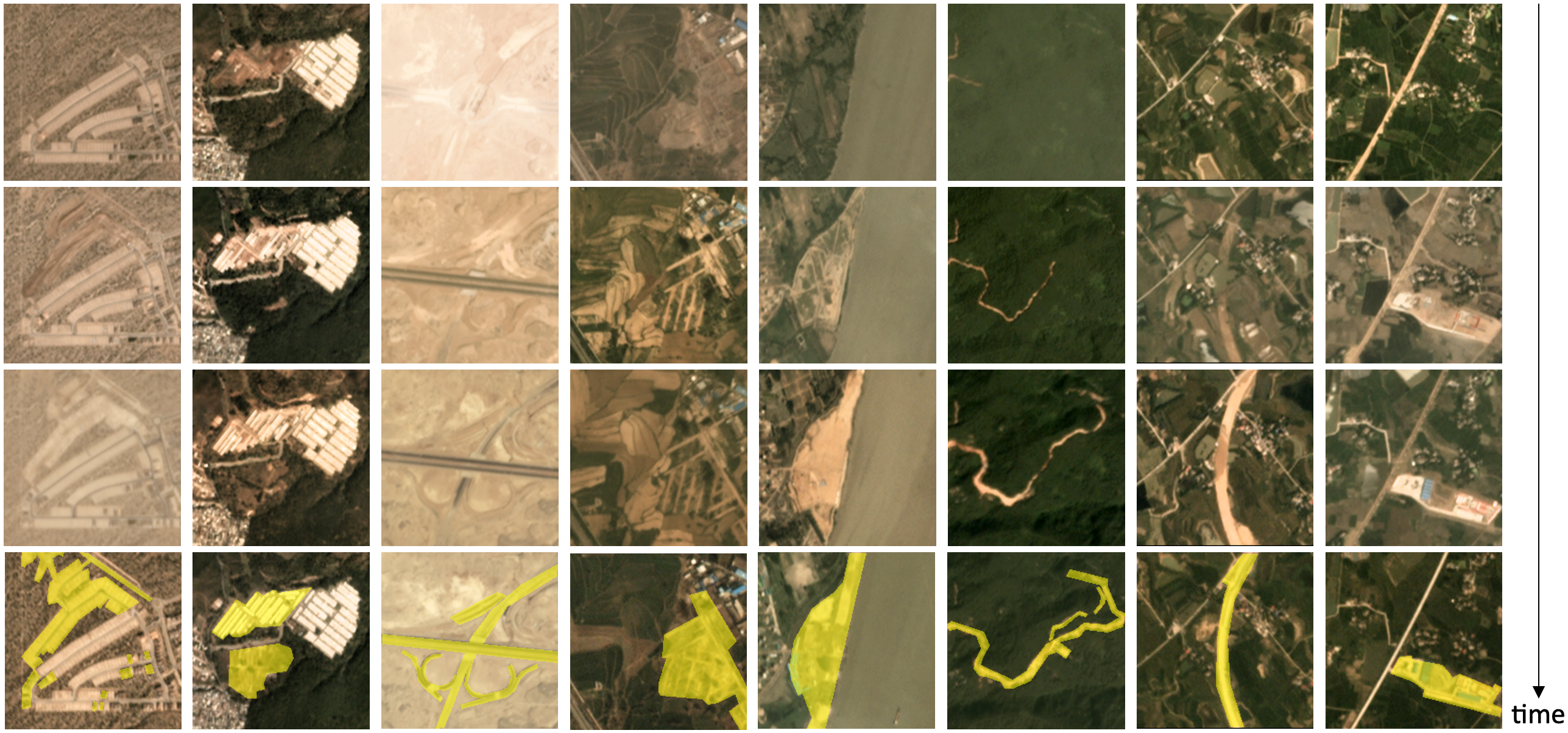}
\caption{\textbf{Monotonic MUDS benchmark.} }
  \label{fig:mmuds}
\end{subfigure}%
\begin{subfigure}{.5\textwidth}
  \centering
  \includegraphics[width=\linewidth]{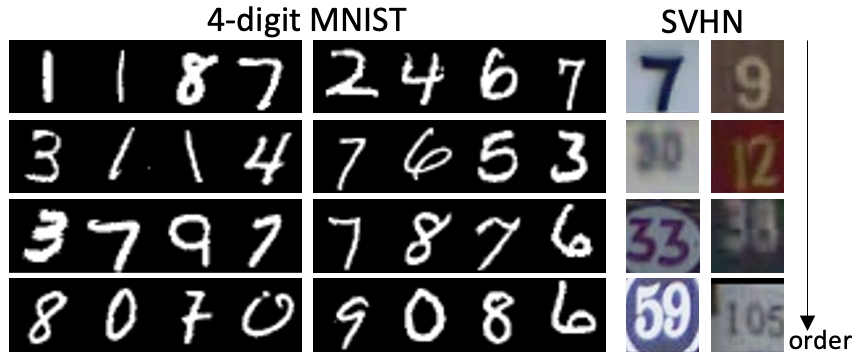}
\caption{\textbf{Image ordering datasets.}   }
  \label{fig:imgorder}
\end{subfigure}

\caption{(a) To evaluate localization and segmentation performance, we manually annotate the monotonically changing regions (shown in yellow) on the MUDS test set. Each sequence contains four frames, and the monotonic changes between the first and last frames are annotated. (b) 4-digit MNIST \cite{lecun1998gradient} (left) and SVHN \cite{netzer2011reading} (right). The task is to order the images by the numbers they contain in increasing order (top to bottom).}
\label{fig:test}
\end{figure}

In addition, 
we showcase our general ordering capability by evaluating our method on standard benchmarks. We compare our ordering performance with related works~\cite{petersen2021differentiable,petersen2022monotonic, grover2019stochastic, cuturi2019differentiable} on the task of sorting images of numbers in ascending order, as shown in Figure \ref{fig:imgorder}. What the model has to learn here is different from the previous datasets, as the ordering is \textit{absolute}, and not understanding changes.

\noindent\textbf{Multi-digit MNIST} \cite{lecun1998gradient} dataset is a modified version of the MNIST dataset in which four digits are concatenated to form a four-digit number. The goal is to order the image sequences in increasing order.
To construct this dataset, we synthetically combine examples from the corresponding train and test sets of the MNIST dataset, resulting in a total of 50,000 training and 10,000 testing images. %\weidi{what is the goal of this dataset ? four images concatenated together ? then do what ?}

\noindent\textbf{Street view house numbers (SVHN)} \cite{netzer2011reading}, was collected from Google Street View and includes images of house numbers. Similarly, the task is to order these numbers in increasing order. The dataset consists of 33,402 images for training and 13,068 for testing.
To ensure consistency with previous studies, we followed the data preprocessing and augmentation methods described in \cite{goodfellow2013multi}.

\subsection{Evaluation metrics}

\noindent\textbf{Localization.} We use a pointing-game evaluation method,
this is to follow the convention of the localization literature in other domains, 
including audio-visual localization~\cite{afouras2020self, arandjelovic2018objects} and saliency methods~\cite{fong2017interpretable, fong2019understanding}, that is, if the pixel with maximum activation in the attribution map contains the change of interest, then it is positive (1), otherwise it is negative (0). As our method only outputs patch-level attribution, we simply select the centre pixel of the patch as the highest activation. 
The overall accuracy is then calculated as the average over all sequences.

\noindent\textbf{Segmentation.}
We use the standard metric for segmentation,
{\em i.e.}, mean intersection over union (mIoU), where the mean is the average across all sequences.

\noindent\textbf{Ordering.} 
We use the evaluation metrics for sorting as outlined in~\cite{petersen2021differentiable, petersen2022monotonic}. 
These metrics include exact match (EM) and elementwise (EW) accuracy. 
EM is considered correct if the entire sequence is ordered correctly, 
while EW considers the order accuracy per element. 
%For instance, if the ground-truth order is $(0,2,1,3)$ and the prediction is $(0,2,3,1)$, the resulting EM would be 0 and the EW would be 0.5. 
Following previous benchmarks we evaluate these at sequence lengths 5, 9 and 16.
To test the generalization to different sequence lengths, 
we also evaluate the exact match accuracy at a fixed sequence length of 5 at test-time (EM5), regardless of the training sequence length. 

\subsection{Implementation details}
We split each dataset into disjoint training and testing subsets,
and then randomly sample frames from each video. 
We keep the time gap between sampled frames constant within each video, but vary this across videos to train a robust model. For image sequences (MUDS, CalFire, OASIS-3) where data collection is less regular, we relax these constraints and simply randomly sample between all frames within the sequence. 
%\charig{This has always been the case, I just thought it's not worth mentioning until it became a ground for rejection...}
We train each model separately for each dataset.

\noindent\textbf{Architecture. } For the encoder, we use a smaller version of TimeSFormer~\cite{timesformer} with a divided space-time attention architecture, consisting of 256 dimensions, 4 heads, 6 layers, and 512 MLP dimensions. For the decoder, we use a standard transformer decoder with the same parameters, except for 64 dimensions and 3 layers. 
As a result, the model is lightweight, with only 4M parameters. 
We use Adam optimizer \cite{kingma2014adam} with learning rate 1e-4 in all experiments, and batch size 32 sequences with 4 frames per sequence for all video datasets, except 3 frames for OASIS-3 as this is the minimum MRI scan sessions per subject. For image ordering, we use batch size 100 with varying numbers of frames. All experiments are run on a single GPU. The code, datasets, and models will be released.

%% file: sec/5_results.tex
\begin{figure*}[t]
  \centering
  \includegraphics[width=\textwidth]{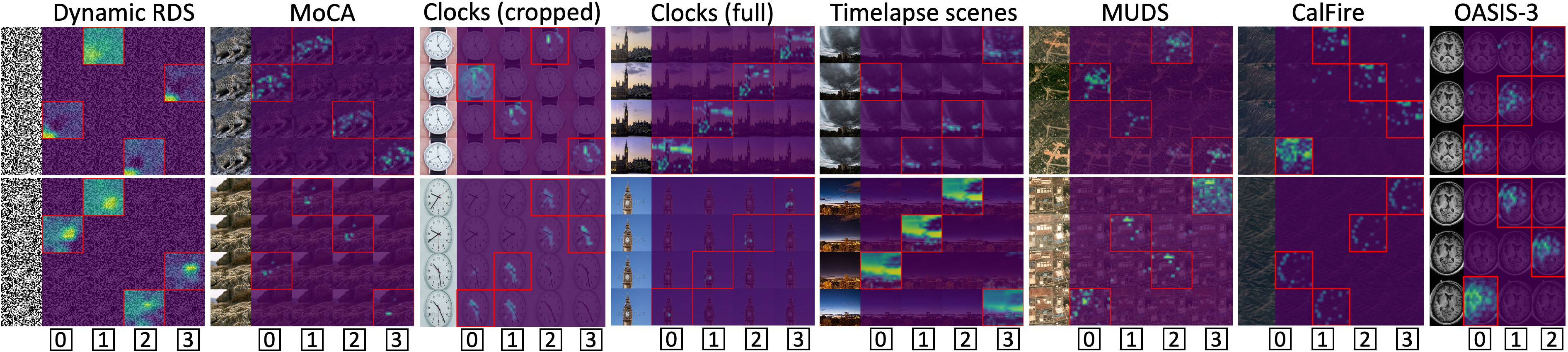}
  \caption{\textbf{Ordering and Localization results} across various datasets, where the model is able to discover and localize various cues across different domains, including object motion, clocks, scenery, landscape and biological aging. 
  The left column shows the input (unordered) images. 
  Each column of the similarity matrix represents the model's prediction of each individual order (0, 1, 2, 3), where the image in the red box is chosen, and the attention heat map within the box localizes the change.
  }
  \label{fig:viz}
\end{figure*}

\section{Results}% 

%We present ordering results  in Section~\ref{sec:results}. Then, as our work cuts across several domains, we consider several avenues for comparison. In Section~\ref{sec:compare_change}, we compare our method with existing change detection methods, and show that discovering monotonic temporal changes is unattainable by previous methods. In Section~\ref{sec:compare_proxy}, we compare with other self-supervised proxy tasks and show that (i) our method works better than previous works in change localization, and (ii) our method learns better representations that can be finetuned to downstream tasks. Lastly, we show in Section~\ref{sec:compare-order} that our model can serve as a general-purpose ordering method that outperforms existing methods in image ordering benchmarks.

\subsection{Results on ordering video frames or image sequence}

\label{sec:results}

\noindent \textbf{Video ordering results.}
The results for ordering as well as the main signals that the model can pick up, are shown by EM and EW in Table~\ref{tab:results}, with qualitative results in Figure \ref{fig:viz}.
The model is able to successfully order sequences across different datasets, particularly in cases where changes are significant. 
It is expected that the scores are not perfect, as sampled data from videos is not guaranteed to contain ordering cues% thus rendering the sequence simply unorderable
, as illustrated in the Supplementary.

\noindent \textbf{Temporal image sequence ordering results.}
For satellite image sequences, the model is able to discover cues that are relevant to ordering, including road building and forest fires. This illustrates our model's application on remote sensing imagery. In MRI scans, we explore the cues for age changes. Our results show that there are cues in the posterior part of the brain.  This is consistent with the literature \cite{blinkouskaya2021brain} that suggest that ventricular enlargement is a prominent feature, and causes the posterior horn to inflate in response to tissue loss. There are also some cues along the outline. 
This is in line with the literature \cite{svennerholm1997changes, scahill2003longitudinal}, which suggest that brain volume also decreases with old age.

\begin{figure}[t]
\centering
\begin{subfigure}{.5\textwidth}
  \centering
  \includegraphics[width=\linewidth]{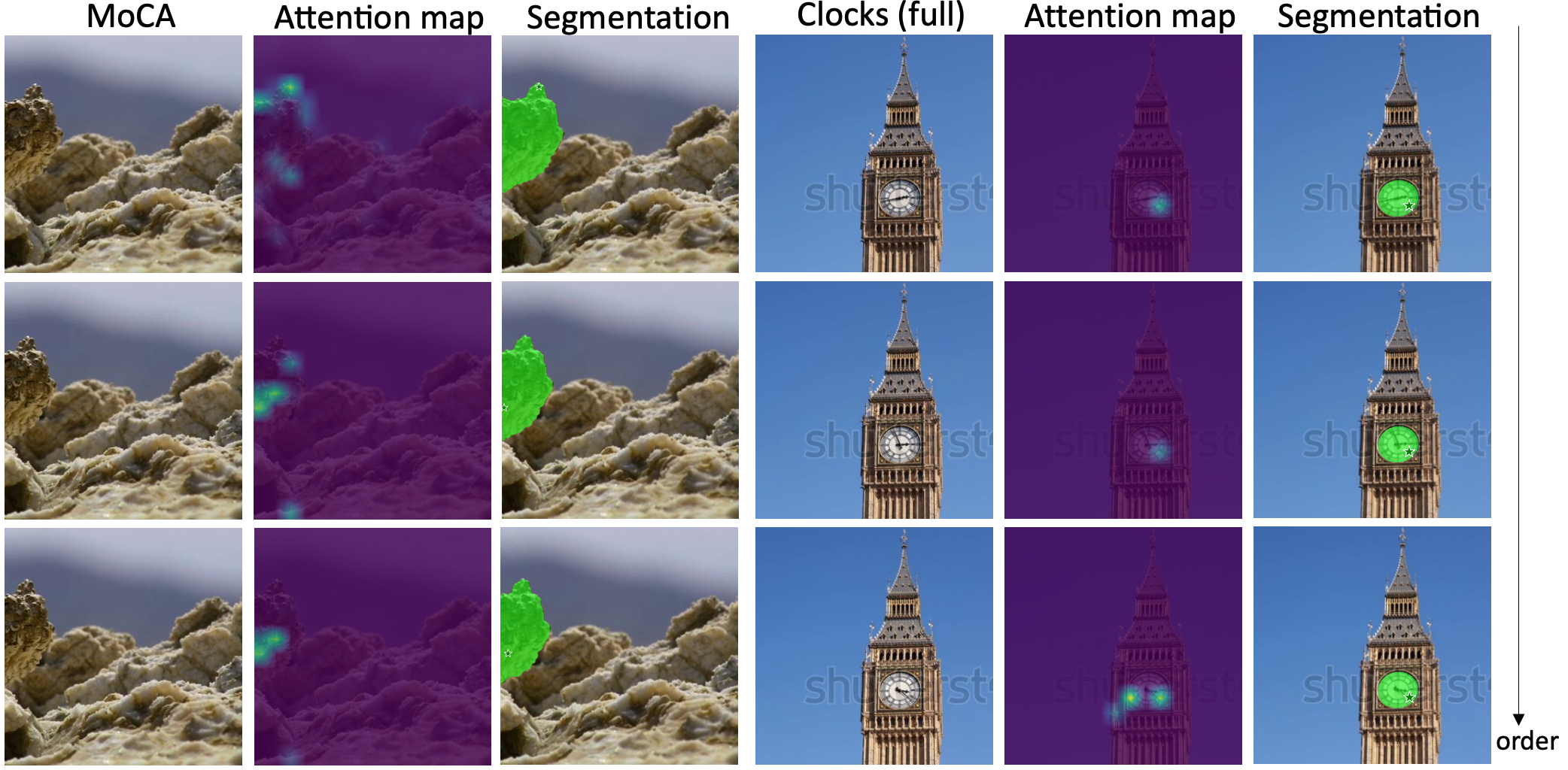}
\caption{\textbf{SAM Prompting.} }
  \label{fig:sam}
\end{subfigure}%
\begin{subfigure}{.5\textwidth}
  \centering
  \includegraphics[width=\linewidth]{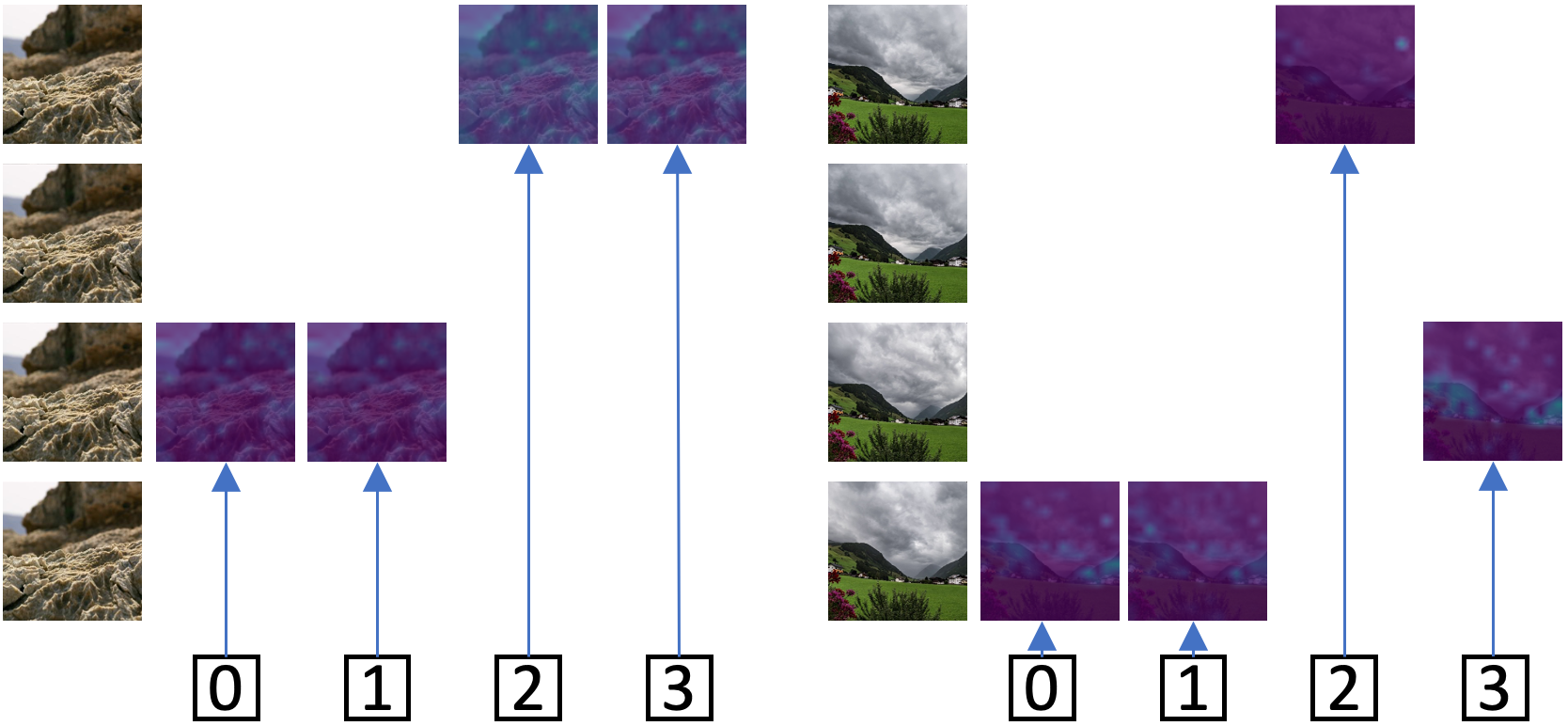}
\caption{\textbf{Failure cases.}   }
  \label{fig:fail}
\end{subfigure}
\caption{(a) The attribution map is used to prompt SAM to obtain segmentation masks. (b) Unorderable sequences, one being too static and the other being too stochastic.}

\label{fig:test}
\end{figure}

 \noindent \textbf{Failure cases.}
A limitation of our model is that we do not force a one-to-one matching between queries and images, 
and this may result in some images being claimed by multiple different queries or by none at all, as seen in Figure~\ref{fig:fail}. 
This problem can easily be resolved by allowing each image to be predicted once. However, not being able to order also provides valuable information as not all real sequences can be ordered -- for example, sequences where everything is static for a period, or very stochastic. Therefore, we treat invalid orderings as a means to provide information on whether particular frames can be ordered or not. {We report further experimental analyses on the failure cases in the Supplementary.}

\subsection{Comparison with change detection methods}
\label{sec:compare_change}

\begin{figure}[t]
  \centering
  \includegraphics[width=\textwidth]{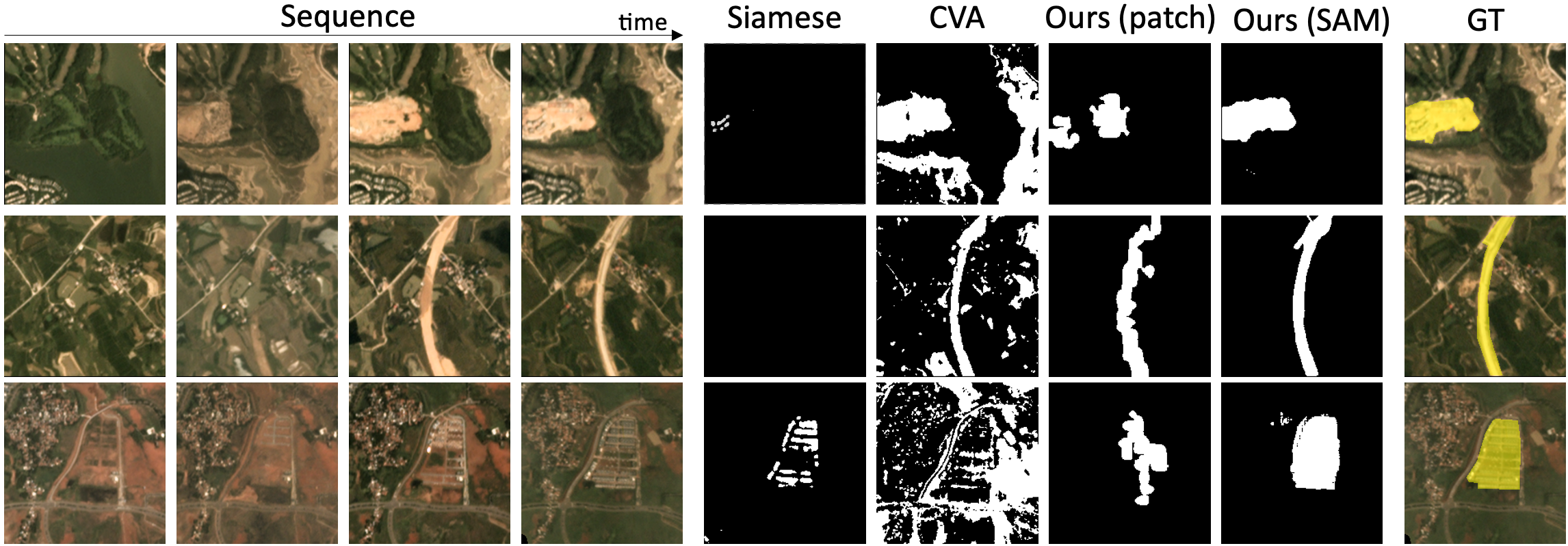}
  \caption{\textbf{Segmentation comparison} with other methods: Siamese networks \cite{hafner2022urban} and CVA \cite{malila1980change}. Supervised methods ignore changes other than building, and pixel-based methods over-segments non-monotonic regions.
  }
  \label{fig:seg_compare}
\end{figure}

\begin{table}[t]
    \centering
\scriptsize
\begin{tabular}{l|cc|c|c} 
\toprule
 & \multicolumn{2}{c|}{Mono-MUDS} & RDS & MoCA \\
Method &
 loc (acc) $\uparrow$ &  seg (mIoU) $\uparrow$ &  seg (mIoU) $\uparrow$ 
&  loc (acc) $\uparrow$ \\ \midrule
Siamese \cite{hafner2022urban}  &         73.3&11.1                    &     --       &    --  \\
CVA  \cite{malila1980change}   &          71.7&34.6                 &     22.3       &  69.6    \\
DCVA \cite{saha2019unsupervised}   &         70.0 & 35.5           &    --     &    --  \\
Ours    &              \textbf{83.3} & {37.9}       &       \textbf{34.2}     &  \textbf{75.0}  \\ 
Ours + DINOv2   &              {80.0} & {41.3}       &      --     &  --  \\ 
Ours + SAM    &              \textbf{83.3} & \textbf{45.1}       &      --     &  --  \\ 

\bottomrule
\end{tabular}
\caption{\textbf{Localization and segmentation results} via the pointing game accuracy for localization, and mIoU for segmentation. The methods for segmentation (patch and SAM) are described in Section \ref{sec:diss}.}
\label{tab:loc}
\end{table}

We compare against three previous approaches for the ability to detect monotonic temporal changes on Monotonic MUDS dataset, 
namely, a supervised Siamese Networks~\cite{hafner2022urban} trained on MUDS dataset for the task of \textit{urban development tracking}, by highlighting the differences between buildings; Change Vector Analysis~\cite{malila1980change}, which is a baseline for the task of \textit{change detection}, highlighting all changes in an image pair without knowledge their nature, and Deep CVA~\cite{saha2019unsupervised}, a learned version of CVA that has been trained specifically on images.

The performance comparison is given in Table~\ref{tab:loc}, and
illustrated in Figure~\ref{fig:seg_compare}. As can be seen, the urban development tracking method~\cite{hafner2022urban} under-segments changes that are correlated with urbanisation, as it is only trained to look at buildings. The change detection methods of~\cite{malila1980change, saha2019unsupervised} over-segments changes that are non-monotonic as it has no concept of time. Both prior methods have shortcomings in detecting urban development: the former method misses changes like road building and deforestation, and the latter includes many erroneous regions such as the seasonal changes in vegetation and water level. Our model is able to highlight correctly the monotonic changes while being invariant to other changes. We further note that our model \textit{discovers} such changes without any prior information on what to look for. {Additionally, replacing the patch projection with DINOv2 and using the localization to prompt SAM both improve the results}. We also evaluate on two other datasets: RDS and MoCA, where we have ground-truth for the moving objects in the video; and show that we obtain favourable results. Note that Siamese and DCVA are only trained on satellite images, hence do not generalize to other domains.

Quantitatively, in Table~\ref{tab:loc}, we find that (i) our pointing-game localization and patch-level segmentation results outperform other methods despite operating only on patch-level $(7 \times 7)$ and not pixel-level granularity, and (ii) segmentation via SAM prompting further improves the results. 

Qualitatively, the results of our localization experiments on various video datasets are presented in Figure~\ref{fig:viz}. The model is capable of accurately identifying monotonic changes while remaining invariant to unrelated cues. Notably, these results were achieved on sequences unseen during training. We additionally show that our localization map works as a good prompt for the Segment Anything (SAM) model to obtain object-level changes, as in Figure~\ref{fig:sam}.

\begin{table}[t]
    \centering
    \scriptsize
    \begin{tabular}{l|cc|c} \toprule
         methods & loc (acc) $\uparrow$ & seg (mIoU) $\uparrow$ & finetuning (F1) $\uparrow$ \\ \midrule
         scratch & --- & --- & 18.2 \\
         S\&L \cite{misra2016shuffle}  & 23.3 & 20.9  & 15.8  \\
         OPN \cite{lee2017unsupervised}  & 25.0 & 24.1 &  17.1 \\ 
         %AoT \cite{wei2018learning}  & \multicolumn{2}{c|}{did not converge} & --- \\
         Ours (AoT proxy)  & 63.3 & 26.9 & 25.5 \\
         Ours & \textbf{83.3} & \textbf{45.1} & \textbf{30.2} \\
        \bottomrule
    \end{tabular}

\caption{\textbf{Comparison with self-supervised methods} on localization and segmentation on Mono-MUDS and on fine-tuning for building change detection on MUDS.
}
\label{tab:compare_Self}

\end{table}

\subsection{Comparison on self-supervised proxy tasks}
\label{sec:compare_proxy}
Here, we compare to other self-supervised methods based on {\em time}. 
We include a discussion in the Supplementary on the subtle differences between the proxy tasks.
We include results for training baselines from scratch on MUDS, 
and testing on Mono-MUDS in localizing and segmenting monotonic changes. 
The results are shown in Table~\ref{tab:compare_Self},
where we conduct three sets of experiments, as detailed below.

First, we observe that previous methods are extremely crude in localization;
this is expected, as they are all based on conv5 feature (even AoT, via CAM). Given a 224p input, conv5 has a $13 \times 13$ feature map with 195p receptive field. 
Our architecture handles localization by design, and is hence more capable than other methods that use post-hoc attribution methods on top of standard backbones.
We also note that AoT does not train, which is also expected as it ingests optical flow as input (and in satellite images using flow does not make sense (change$\neq$motion)), our method is more flexible in this regard.

Second, we ask if our method is still superior if the architectural gap is closed. To achieve this, we also compare the proxy task in AoT (time direction) with ours (ordering) under a fairer comparison (using our architecture and RGB input), and show this in the table under
``Ours (AoT proxy)''. We conclude that both our architecture and proxy task leads to significant improvements.

Third, we investigate fine-tuning on a target task that requires both time and localization: change detection of buildings (like \cite{hafner2022urban}). For each method, we pre-train the encoder on MUDS, freeze it, then train a lightweight head ($3\times$ deconvolutions) on top of the same dataset, and test on unseen sequences. To keep this fair, we keep the number of parameters roughly the same across methods. The results (Table \ref{tab:compare_Self}, right column) show that our method learns good representations as compared to previous self-supervised methods in localization tasks.

\subsection{Comparison with image ordering methods}
\label{sec:compare-order}

We compare to two previous methods on image ordering benchmarks where the task is to arrange the images in increasing order. {\em Differentiable sorting networks} such as DiffSort~\cite{petersen2021differentiable} and its successor 
DSortv2~\cite{petersen2022monotonic} employ a parameter-free sorting network to rank scalars. 
%We compare with these sorting models, and show that our model is capable as a general ordering network with added functionality.
{\em Pointer Networks} ~\cite{vinyals2015pointer}
ranks features by using a recurrent encoder-decoder network with attention. We note that Ptr-Net is not initially designed for such a task, but for arranging a set of coordinates. We simply extend pointer networks by adding an image encoder and task the model to rank the image features from small to large. We then jointly train this encoder and the pointer network. 
For fair comparison, we use the same transformer encoder as our model, 
and use the pointer network as the decoder with similar size to our transformer decoder. 

The quantitative results are presented in Table~\ref{tab:compare_results}.
Our results demonstrate that (i) we compare favourably in ordering performance -- on both MNIST and SVHN, our method has the best performance of the four (ii) Ptr-Net does not automatically generalise to testing with different sequence lengths, while our method does (as reflected by the poor EM5 accuracy), and (iii) our method also has the added benefit of having an attribution map.

\begin{table}[t]
\scriptsize
\setlength{\tabcolsep}{2pt}
\begin{tabular}{l|ccc|ccc}

\toprule
  dataset      & \multicolumn{3}{c|}{MNIST} & \multicolumn{3}{c}{SVHN} \\ \midrule
frames   & 5     & 9    & 16    & 5    & 9    & 16   \\ \midrule
DSort \cite{petersen2021differentiable}            & 83.4|92.6 & 56.3|86.7 & 30.5|80.7|86.6 & 64.1|82.1 & 24.2|69.6 & 3.9|59.6|66.8  \\
DSv2 \cite{petersen2022monotonic} & 84.9|93.2 & 63.8|89.1 & 31.1|82.2|----- & 68.5|84.1 & 39.9|75.8 & 12.2|65.6|----- \\ %\midrule
Ptr-Net \cite{vinyals2015pointer} &     91.9|95.6    &  87.7|95.0      &  68.9|90.0|1.1  &   76.3|87.6     &  48.7|79.4      &  9.8|63.2|0.1      \\
Ours    &    \textbf{93.9}|\textbf{96.7}     &    \textbf{87.9}|\textbf{95.2}    &    \textbf{72.2}|\textbf{91.2}|\textbf{92.9}     &     \textbf{77.3}|\textbf{88.2}   &  \textbf{53.9}|\textbf{81.0}      &  \textbf{19.4}|\textbf{67.9}|\textbf{67.6}   \\ \bottomrule

\end{tabular}
\caption{\textbf{Ordering on image datasets} on two standard benchmarks (MNIST and SVHN) where the task is to order images of numbers in increasing order. Metrics are (EM|EW|EM5). EM and EW are evaluated at the sequence length the model has been trained on (5/9/16), whereas EM5 tests generalisation to test length 5.
}
\label{tab:compare_results}
\end{table}

%% file: sec/6_conclusion.tex
\section{Conclusion}
In this paper, we explore using time as a proxy loss for self-supervised training of models to discover and localize monotonic temporal changes in image sequences. Possible extensions include discovering more complex temporal changes (seasonal/periodic), or object state and attribute changes. 
It would also be interesting to investigate how the model scales with larger datasets and compute, and what new applications this task can enable. Overall, we hope this paper presents a valuable starting point for future research and applications in this area.

\section*{Acknowledgements} 
We thank Tengda Han, Ragav Sachdeva, and Aleksandar Shtedritski for suggestions and proofreading.
This research is supported by the UK EPSRC CDT in AIMS (EP/S024050/1),
and the UK EPSRC Programme Grant Visual AI (EP/T028572/1).

%%%%%%%%%%%%%%%%%%%%%%%%%%%%%%%%%%%%%%%%%%%%%%%%%%%%%%%%%%%%

%% file: sec/X_suppl.tex
%\clearpage
\setcounter{page}{1}
%\onecolumn
\appendix

%\clearpage
%\tableofcontents
%\weidi{do a content page to make it easy to read ?}

\noindent The Supplementary Material includes the following sections:
\begin{itemize}
    \item \textbf{Sect.~\ref{sec:supp1_data}: Datasets in more detail}, where the datasets used in the paper are explained further.
    \item \textbf{Sect.~\ref{sec:supp2_code}: Code}, where we present the pseudocode alongside the attached code.
    \item \textbf{Sect.~\ref{sec:supp3_related}: Related work in more detail}, where we compare our method against previous ones and highligh the differences.
    \item \textbf{Sect.~\ref{sec:supp4_failure}: Unorderable sequences}, where we discuss the model's handling of unorderable sequences. %\az{I changed this from Failure mode}
    \item \textbf{Sect.~\ref{sec:supp5_settings}: Experimental settings}, where we perform variations on the train/test set division.
    \item \textbf{Sect.~\ref{sec:supp6_ablation}: Ablation studies}, where we perform variations on the architecture.
    \item \textbf{Sect.~\ref{sec:supp7_qualitative}: Qualitative results}, for both orderable and unorderable cases.
\end{itemize}

\section{Datasets in more detail}
\label{sec:supp1_data}

Table \ref{tab:data} expands on the main paper's dataset attributes (Main paper, Table 1) to give more detailed dataset statistics. 

\begin{table}[H]
\small
\centering
\setlength\tabcolsep{1pt}
\begin{tabular}{lccc|ccc}
\toprule dataset & resolution & patch size &  \# patches  & \# train & \# test  \\ \midrule
MNIST  & 28$\times$112 & 7 & 4$\times$16  &  50000 & 10000 \\ % & --- & --- \\ 
SVHN & 54$\times$54  & 6 & 9$\times$9  &  33402 & 13068 \\
\midrule
Dynamic RDS  & 42$\times$42 & 7 & 6$\times$6  &  $\infty$ & $\infty$ \\ % & --- & --- \\ 
MoCA & 336$\times$336  & 21 & 16$\times$16  &  75 & 13\\
Clocks (cropped) & 196$\times$196  & 14 & 14$\times$14 & 2011  & 500 \\
Clocks (full) & 320$\times$480  & 20 & 16$\times$24  &  210 & 50 \\
Timelapse scenes& 336$\times$336   & 21 & 16$\times$16 &  130 & 50\\
MUDS & 196$\times$196  & 7 & 28$\times$28  &  60 &  20 \\
CalFire & 336$\times$336  & 21 & 16$\times$16  &  800 &  276 \\
OASIS-3 & 224$\times$224  & 16 & 14$\times$14  &  50 &  19 \\
 \bottomrule
\end{tabular}
\vspace{4pt}
\caption{\textbf{Dataset attributes.} The different video datasets used, and their attributes in detail.}
\label{tab:data}
\end{table}

\newpage
\section{Code}
\label{sec:supp2_code}

The pseudocode is shown below, with the full code being available on the project webpage.

\input{suppfig/pseudocode}

\section{Related work in more detail}

\label{sec:supp3_related}

%This section outlines more explicitly the similarities and differences between our work and related works.

\subsection{Self-supervised learning from time}

\begin{table*}[h]
\scriptsize
\begin{tabular}{l|llll}
\toprule
              & S\&L \cite{misra2016shuffle}             & OPN   \cite{lee2017unsupervised}           & AoT  \cite{wei2018learning}              & Ours                \\ \midrule
Input         & RGB              & RGB              & Flow               & RGB                 \\
Filtering     & Flow             & Flow             & Flow               & no                 \\
Evidence      & post-hoc (pool5) & post-hoc (pool5) & post-hoc (CAM)     & built-in            \\
Training task & is middle frame in-between & shuffle and order & forward or backward & shuffle and order \\
Goal          & rep. learning    & rep. learning    & rep. learning+apps & change localization 
   \\ \bottomrule
\end{tabular}

\caption{\textbf{Comparison to related works}. This table compares this paper with several related works, namely Shuffle and Learn \cite{misra2016shuffle}, Order Prediction Network \cite{lee2017unsupervised}, and Arrow of Time \cite{wei2018learning}. Unlike previous work where the goal was
self-supervised representation learning, our goal is to directly discover the change and predicts its localization.}
\label{tab:rel-works}
\end{table*}

\vspace{5pt}\noindent \textbf{On motivation.}
The main goal of Shuffle and Learn \cite{misra2016shuffle} and extensions such as Order Prediction Network \cite{lee2017unsupervised} is representation learning using self-supervision as pre-training – the authors had the intuition that change was required and used a filter for this based on optical flow to select suitable frame-sets in order to learn good representations. Arrow of Time \cite{wei2018learning}, too, focuses substantially on representation learning, though based on optical flow as input, and they too filter away sequences that lack motion using optical flow.

\vspace{5pt}\noindent \textbf{On attribution learning.}
All existing methods apply attribution methods post-hoc (S\&L and OPN look at the activation on the spatial pooling after conv5 and AoT uses CAM on conv5 features).   As mentioned in the main paper, both of these post-hoc methods have low accuracy as they
%\az{which paper are you talking about here? or does this apply to all citations?} 
rely on 7x7 conv5 features with very large receptive field. In contrast, we make a contribution in designing an architecture that has attribution built-in, enabling more precise attribution (please refer to the respective figures in each paper -- Fig. 4 of \cite{misra2016shuffle}, Fig. 8 of \cite{lee2017unsupervised} and Fig. 4 of \cite{wei2018learning}), and no need for post-processing.

\vspace{5pt}\noindent \textbf{On pretext task.}
AoT asks the model to distinguish between forward and reverse order of an otherwise in-order sequence, while S\&L fixes the start and end frames and asks if the middle frame is inside (see Sec. 3.2 of their paper for details). For our task it does not matter if the sequence is forwards or backwards -- but that is the only goal of AoT, and the AoT goal would not be possible with shuffled frames. Our task is identical to OPN where the entire sequence is shuffled and the ordering is predicted, but we add (i) built-in attribution, and (ii) a more flexible architecture capable of general ordering across a variable number of frames. Without all these three add ons, our work would essentially boil down to the same as that of OPN.

\subsection{Ordering methods}

\begin{figure}[h]
\begin{floatrow}
\ffigbox{%
  \includegraphics[width=.5\textwidth]{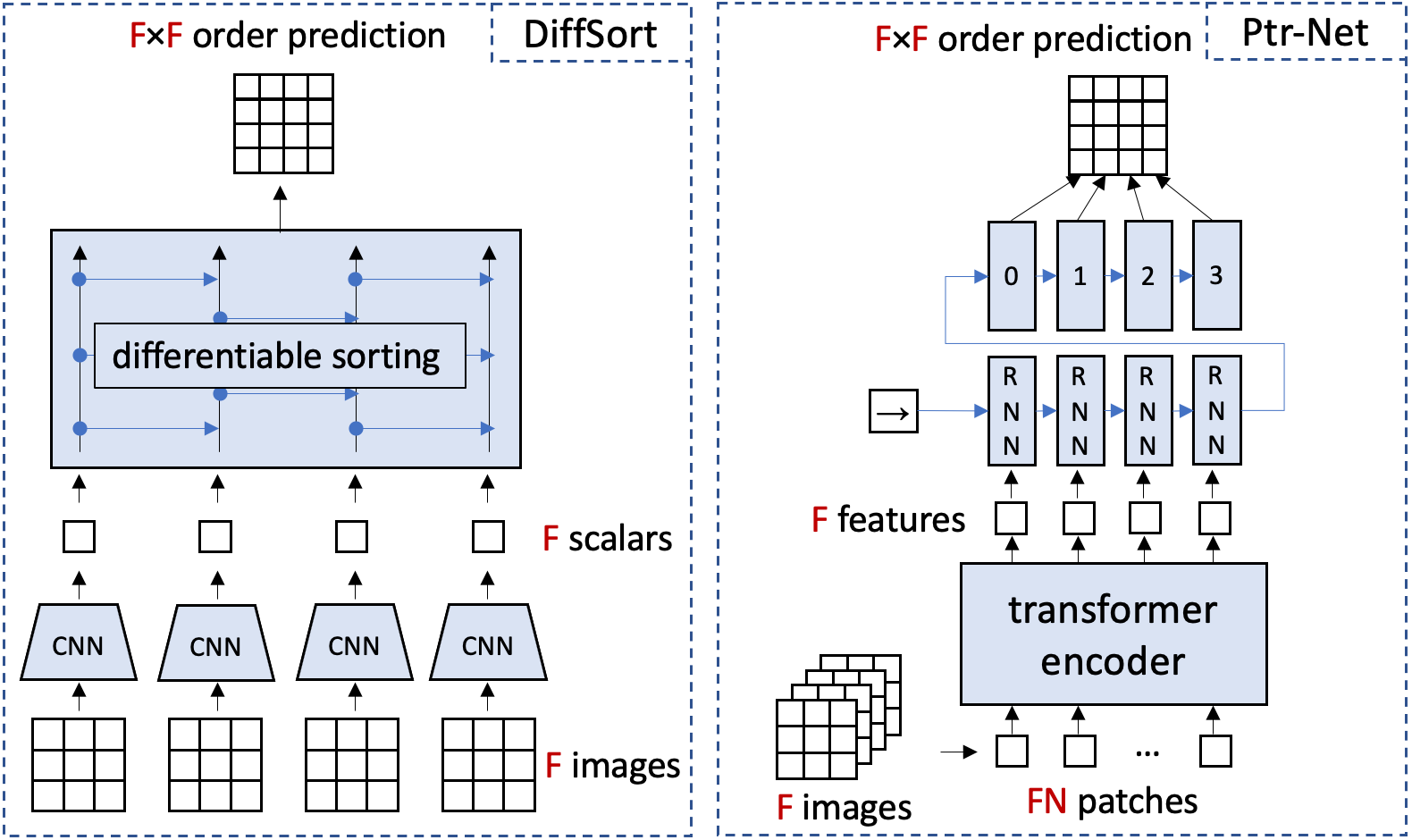}%
}{%
  \caption{\textbf{Architectures of related works} for ordering, including Differentiable Sorting \cite{petersen2021differentiable} and Pointer Networks \cite{vinyals2015pointer}.}%
  \label{fig:related}
}

\capbtabbox{%
  
  \setlength\tabcolsep{5pt}
  \begin{tabular}{l|ccc}
    \toprule  
    & DSort & Ptr & Ours  \\ 
    \midrule
    Absolute order & $\checkmark$ & $\checkmark$ & $\checkmark$  \\
    Relative order & \xmark  & $\checkmark$ & $\checkmark$ \\ 
    Flexible length & $\checkmark$  & \xmark & $\checkmark$ \\ 
    Localization & \xmark & \xmark & $\checkmark$ \\ 
    \bottomrule
  \end{tabular}
}
{%
\caption{\textbf{Related works.} We show DiffSort \cite{petersen2021differentiable}, and Pointer Networks \cite{vinyals2015pointer} are less versatile.}
  \label{tab:comparison}
}
\end{floatrow}
\end{figure}

The other ordering architectures are shown in Figure~\ref{fig:related}, and their comparison is shown in Table~\ref{tab:comparison}. Our architecture allows for all of (i) ordering of absolute cues (e.g.\ images of numbers) (ii) ordering of relative cues (e.g.\ in natural images), (iii) ordering with flexible length during training and testing, and (iv) explicit localization via attribution map. Note that Ptr-Net is not initially designed for such a task, and we extend the architecture while preserving fair comparison.
%\weidi{can we align the figure and table based based on top point?}

%\caption{\textbf{Functionality Comparison.} We compare the functionality of different methods that allow for ordering.}
%\label{tab:compare}

\section{Unorderable sequences}
\label{sec:supp4_failure}

%\az{I think we should rename this as `unorderable frame sets', since a sequence is ordered by definition} \charig{I am hesitant about this (1) Main paper used it (2) I think it might be ok as a sequence is in \textit{an} order by definition, but doesn't mean the order is defined or relevant, e.g. a list ['dog', 'cat', 'charig', '40']. In our context they are timestamped but not chronological (3) I tried and everything sounded more confusing. I have clarified a bit though.}

In real-world scenarios, sequences may not contain monotonic ordering cues for every frame (such as sequences that are static over several frames, or ones where changes are cyclic), and hence it is impossible to order such sequences using monotonic changes alone 
% without access to the original timestamps 
-- we denote such sequences as \textbf{unorderable}. We also observe that the model sometimes makes predictions that are \textbf{inconsistent} %az{can we use inconsistent instead of invalid?} {\color{red} 
by predicting some ordering indices multiple times and others not at all. This happens when %some predicted ordering indices are skipped 
%\az{not clear what `predicted ordering indices are skipped' means} or repeated
%not one-to-one between input and output \az{define input \& output} 
some frames are being claimed by multiple queries and others claimed by none at all. Here, we investigate the correlation between these two occurrences and ask whether the model is able to correctly identify unorderable sequences by making inconsistent predictions. 

To do this, we create an unorderable set by including static frames on Mono-MUDS datset, resulting in 60 unorderable sequences (in addition to the 60 orderable ones) of 4 frames each. We experiment with these 120 sequences, and report the results in Table~\ref{tab:confusion}. We show that there is a strong correlation between the model predicting inconsistent ordering and the sequence being unorderable. We also note that this occurs as a result of a design during inference, hence training is unaffected.

In summary: an inconsistent prediction implies that the set of frames cannot be ordered, and the model is able to flag such sequences. %in practice, such as in filtering away satellite image sequences that lack significant events. \az{this example is not clear -- there may be significant events away from static frames} \charig{not sure what to say instead -- to discuss}
%In practice, knowing that a sequence is unorderable can be beneficial \az{Need to explain this benefit more fully. Or do you mean that an invalid prediction is a benefit, because it implies that the set of frames can't be ordered.} \charig{I initially meant the former, but also included your suggested point}
In the case of videos, the ground-truth ordering is known, so we are able to easily detect when the prediction is incorrect or inconsistent. It is also possible to look at the (lack of) attention in the attribution map to observe the lack of ordering cues. We also show qualitative examples of these cases in Section~\ref{sec:supp7_qualitative2}.

%\az{this section should reference or be merged with the examples in section~\ref{sec:supp7_qualitative}.2}

\begin{table}[h]
    \centering
    \small
    \begin{tabular}{l|ccc} \toprule
         & correct  & incorrect & inconsistent \\ \midrule
        gt orderable & 88.3 & 10.0 $\uparrow$ & 1.7 $\downarrow$\\
        gt unorderable & 0.0 & 1.7 $\downarrow$ & 98.3 $\uparrow$ \\ \bottomrule
    \end{tabular}
    \caption{\textbf{Unorderable sequences.} For an image sequence (0,1,2,3), the model either makes correct (0,1,2,3), incorrect \eg~(0,1,3,2), or inconsistent \eg~(0,1,3,3) predictions. We investigate the correlation between the model making inconsistent predictions and the sequence being unorderable. The model reliably gives inconsistent predictions for unorderable sequences.}%  \az{ you need to explain what the column headings mean here}} %\az{not clear why this is a confusion matrix}
    \label{tab:confusion}
\end{table}

\section{Experimental settings}
\label{sec:supp5_settings}

\begin{figure}[h]
  \centering
  \includegraphics[width=.95\textwidth]{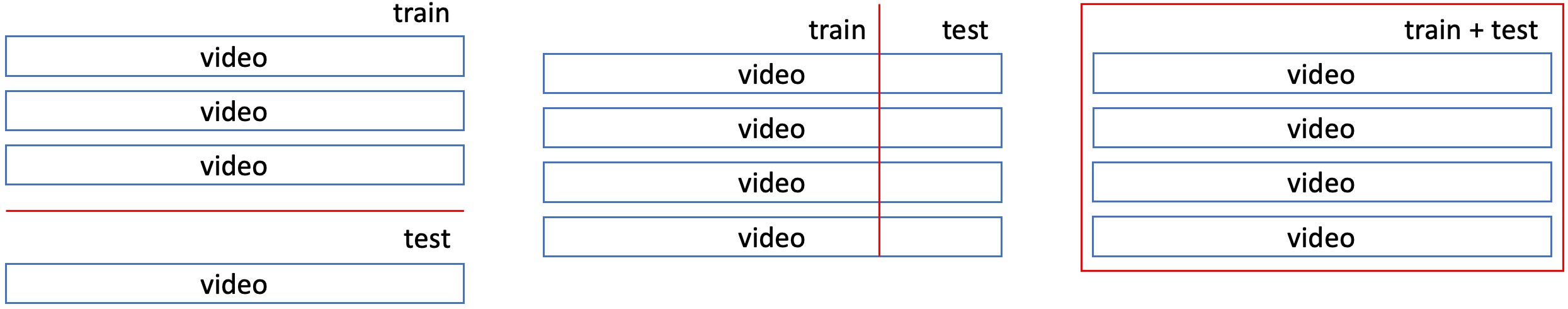}
  \caption{\textbf{Experimental settings.} The main paper only explores the case where the train and test videos are separate videos (left), all of the same class. Here we  explore two other settings: testing on the same clips but at different time (centre), and testing on the training set (right).
  }
  \label{fig:setting}
\end{figure}

In the main paper, we only explore the setting where the train and test sets are independent video clips within the same dataset, i.e. Figure \ref{fig:setting}, left. However, as a self-supervised task, we can explore other settings that are appropriate for applications.

\noindent{\textbf{Testing on the same clips, but at different times}} (Figure \ref{fig:setting}, centre). There may be cases in remote monitoring and surveillance where there exists past data that can be used for training, and the adaptation is mostly towards new sequences that are taken under the same or similar settings. In such cases, we can split the training and testing sets by time instead of by video clips. %For this, we experiment with MUDS dataset, splitting the sequences by time instead of videos. For each sequence of 24 frames, we use the first 16 frames for training and the last 8 for testing.

\noindent{\textbf{Testing on the training set}} (Figure \ref{fig:setting}, right).
Since no annotations are used, 
we can also test directly on the training set as an exploration tool to investigate what cues are used in determining the order. 
%We find that overfitting to too little data (i.e. a single video) easily allows overfitting, \weidi{rephrase the sentence}
%so we instead perform inference on the training set of the datasets. 
%To keep fair comparison we use the original train/test split and do not train on the test set.

\vspace{10pt}\noindent{\textbf{Experiment.}} We conduct an experiment by having a common test set, and perform training on three different training settings, namely: (i) the training set does not include any videos from the test set, (ii) the training set includes the videos from the test set, but they do not overlap in time, and (iii) the training set includes the entire testing set. To keep the experiments fair we use the same number of clips for each training setting. We test this on MUDS dataset, and show the resutls in Table \ref{tab:traintest}. It shows that the accuracy increases as the domain gap between training and test sets decreases.

\begin{table}[H]
\small
\centering
\setlength\tabcolsep{10pt}
\begin{tabular}{l|cc}
\toprule setting & EM↑ & EW↑  \\ \midrule
Base (split by video) & 45.1  &  52.1  \\
Testing on same clips, different time & 49.7  & 59.6  \\
Testing on training set &  88.5 & 91.3   \\
 \bottomrule
\end{tabular}
\vspace{4pt}
\caption{\textbf{Experiment on different settings.} This table shows the accuracy for the common test set in MUDS dataset. Metrics are exact match and elementwise accuracies (higher is better).}
\label{tab:traintest}
\end{table}

\section{Ablation studies}
\label{sec:supp6_ablation}

In this section, we conduct ablation experiments on different aspects of the architectural design. As we introduce a new model in order to solve a new task, there are many variables that are interesting to investigate.
%including on the number of layers, dimensions, and spatial map size. 
%\az{we should not say the following ...}
%We did not tune any of these parameters when designing the model and experiments, and simply chose a reasonable number.
%\weidi{all experiments are on SVHN ?}

\subsection{Transformer Architecture} 
\vspace{-0.1cm}

To investigate the effect of model size on the ordering performance, we experiment by varying the sizes of the encoder and decoder. We do this by varying the numbers of (layers/feature dim/heads/feedforward dim) or both the encoder and decoder (in PyTorch's nn.Transformer, they are referred to as (num\_layers/d\_model/nhead/dim\_feedforward). Feature dimensions refers to the number of features in self-attention or cross-attention within each transformer layer, while the feedforward dimension refers to the dimensionality of the feedforward network applied after attention within each transformer layer.
%\az{clarify what dimensions and `MLP dimensions' refers to here. Also, how many layers is the MLP} 
%of both the encoder and decoder.

\begin{table*}[h]
\scriptsize
\centering
\setlength\tabcolsep{2pt}
\begin{tabular}{lccc|ccc}
\toprule experiment & Encoder & Decoder & \# params & EM↑ & EW↑ & EM5↑  \\ \midrule
Base (small)  & 6/256/4/512 & 3/64/4/512 & 3.6M & 53.9 & 81.0 & 78.0 \\ \midrule
increase encoder size  & 2x base & base & 19.4M &  62.9 & 85.1 & 83.8 \\
increase decoder size  & base & 2x base & 6.4M &  53.9 & 81.5 & 78.7 \\
increase both  & 2x base & 2x base & 22.2M & 62.7 & 84.9 & 83.0 \\ \midrule
decrease encoder size  & 0.5x base & base & 816k &  27.2 & 68.8 & 62.9  \\
decrease decoder size  & base & 0.5x base & 3.3M &  51.7 & 80.1 & 76.8 \\
decrease both  & 0.5x base & 0.5x base & 515k & 25.6 & 68.0 & 61.8  \\
 \bottomrule
\end{tabular}
\vspace{4pt}
\caption{\textbf{Ablation studies.} We perform ablation studies varying the architecture of our model. The numbers for the encoder/decoder are (layers/feature dim/heads/feedforward dim). All ablations are relative to the base size, for example, 2x base refers to (12/512/8/1024). We evaluate on SVHN dataset with 9 frames. Base corresponds to the version implemented in the main paper.}
\label{tab:ablatesupp}

\end{table*}

The results are shown in Table \ref{tab:ablatesupp}. We can see from the results that the encoder size has a significant impact on the accuracy, while the decoder size does not matter as much. This is interesting, and we think that this arises from the fact that the encoder is doing most of the work, both in feature extraction and in ordering. We further investigate this in Section \ref{sec:att}. %This agrees with our intuition -- the encoder is doing most of the work by trying to output the correct ordering for each of the patches, while the decoder simply has to align the query vectors in order towards the correct output. \az{Where does this intuition come from? You need to motivate more why this is so -- possibly give a forward ref to the attention investigation below}
\vspace{-0.2cm}

\subsection{Spatial feature map size} 
\vspace{-0.1cm}
We also experiment varying the spatial feature map size, to look at the trade-off between the granularity of localization and ordering performance. Small patch sizes require significantly higher memory consumption.

\begin{table*}[h]
\scriptsize
\centering
\setlength\tabcolsep{5pt}
\begin{tabular}{lcc|ccc}
\toprule experiment & patch size &  \# patches & EM↑ & EW↑ & EM5↑  \\ \midrule
Base (M)  & 6 & 9$\times$9 & 53.9 & 81.0 & 78.0 \\
S  & 3 & 18$\times$18& 47.5 & 78.2 & 74.9 \\
L  & 9 & 6$\times$6 & 54.4 & 81.5 & 79.8 \\
XL  & 18 & 3$\times$3 & 44.1 & 76.1 & 73.1 \\
 \bottomrule
\end{tabular}
\vspace{4pt}
\caption{\textbf{Patch size ablation.} We perform ablation studies varying the patch size. We evaluate on SVHN dataset with 9 frames. Base is used in the main paper.} %\az{Probably can't do anything about this now, but you are changing patch size and number of patches at the same time. It would be better to vary these independently}} i suppose you mean by also scaling the size of images
\label{tab:ablate2.6}
\end{table*}

As shown in Table \ref{tab:ablate2.6}, while small patches are generally good for visualisation, as the changes will be more finely localized, it does come with a trade-off that ordering accuracy drops when the patches are too small.

\vspace{-0.2cm}

\subsection{Encoder attention} 
\vspace{-0.1cm}

\label{sec:att}
We also look at the choice of attention used in the encoder. 
In particular, in the model of the main paper we use divided space-time attention in the transformer encoder so that tokens of the same spatial position between frames are allowed to communicate.

The benefit of allowing communication between the frames in the encoder is that the ordering can be done within the encoder itself, easing the load on the decoder. Without this, the encoder will merely act as a feature extractor, and the decoder will have to find a way to rank these features. Another benefit particularly in video ordering is that the patches with the same spatial position across the frames are able to communicate and compare with each other, highlighting the differences more easily.

In this section we also experiment with two other settings, 
where (i) attention is restricted to within each frame, and
(ii) all tokens are allowed to communicate with each other (we refer the reader to Figure~2 of \cite{timesformer} for illustration). 
We evaluate both on image set ordering (SVHN) and video temporal ordering (MUDS).

\begin{table*}[h]
\small
\centering
\setlength\tabcolsep{10pt}
\begin{tabular}{lc|ccc}
\toprule experiment & GPU mem (GB)↓ & SVHN↑ & MUDS↑   \\ \midrule
Base (divided space-time) & 8.0/5.3 & 53.9 | 81.0 & 56.4 | 69.6  \\
Space only & 8.1/5.7 & 55.8 | 82.1 & 13.0 | 35.8  \\
Full attention & 14.8/9.2 & 51.5 | 80.6 & 7.6 | 30.7 \\
 \bottomrule
\end{tabular}
\vspace{4pt}
\caption{\textbf{Encoder attention ablation.} We perform ablation studies varying the encoder attention. We evaluate on SVHN and MUDS with 9 and 4 frames respectively. Metrics are (EM | EW). We also show the GPU memory usage in each respective dataset (SVHN/MUDS), and show that full attention significantly consumes more memory.}% \az{explain what the two numbers in the GPU column mean}}
\label{tab:ablate2.5}
\end{table*}

As shown in Table \ref{tab:ablate2.5}, we observe that the results vary significantly with the datasets. In the case of image set ordering on SVHN, inter-frame communication matters less presumably because there is no differences between patches of the same spatial position to highlight, unlike the case of video temporal ordering. We also observe that space-only attention (6 spatial layers) obtains slightly higher accuracy than that of divided space-time attention (3 spatial layers, 3 temporal layers), which reinforces the hypothesis that spatial attention within the image is more important than across images.

In the case of ordering video frames (MUDS), we found that (i) space-only attention hinders the model's ability to communicate with each other, and (ii) full attention is significantly more difficult to train due to a significantly larger number of tokens to attend to. 
%Moreover, having more spatial attention layers (6 vs 3) allows the model to improve the accuracy. %\az{where is the result comparing the number of spatial attention layers?} \charig{what I meant was that if you do 6 spatial layers (space only) it's better than divided (3 space 3 time) because the time layers is not useful here. That's why space-only is a bit better here.}
%divided space-time not only performs better, but also trains significantly faster. We think that the restricted attention allows the model to focus on extracting the spatially aligned changes.
\vspace{-0.2cm}
\subsection{Number of frames}
\vspace{-0.1cm}

In this section, we investigate the trade-off for increasing the number of frames in a sequence. As shown in Table~\ref{tab:ablate2.6}, having more frames allows the cues to be highlighted more clearly, and allows the model to learn better what changes are correlated with time and what are not. However, the number of possible orders will also increase combinatorially which makes the task more difficult. 

\begin{table}[h]
\small
\centering
\setlength\tabcolsep{10pt}
\begin{tabular}{l|ccc}
\toprule experiment & EM↑ & EW↑   \\ \midrule
Base (4)  & 62.5 & 73.0  \\
2  & 51.3  & 51.3   \\
3  & 65.2 &  73.1 \\
5  & 37.1 & 55.0  \\

 \bottomrule
\end{tabular}
\vspace{4pt}
\caption{\textbf{Frame count ablation.} We perform ablation studies varying the number of frames. We evaluate on Timelapse clocks (cropped).}
\label{tab:ablate2.6}
\end{table}
\vspace{-0.2cm}

\subsection{Choice of objective function}
\vspace{-0.1cm}

We have two different loss functions (forward and reversible). Consider a scene of a sky where the sun is moving upwards. Without accounting for reversibility, it is difficult to distinguish between a sunrise and a sunset in reverse, hence unidirectional ordering becomes challenging, sometimes impossible. Since our goal is only to attribute the change, not to determine the forward or backwards direction, our intuition is that considering reversibility in the sequence allows the model to be trained better, as this prevents giving the model confusing training signals. In this case we find that the model is (i) harder to train and (ii) susceptible to overfitting if we do not allow reversibility in the model, especially in datasets where changes are reversible.

Table \ref{tab:moca_ablate} experimental results for the MoCA dataset as an ablation. From the table, the results show that using reversible losses helps with both (i) training and (ii) generalisation.

\begin{table}[h]
\small
\setlength\tabcolsep{5pt}
    \centering
    \begin{tabular}{l|c|cc|cc}
    \toprule
        reversible? & loss↓ & train EM↑ & train EW↑ & val EM↑ & val EW↑ \\ \midrule
         yes & 0.119 & 84.2 & 91.5  & 82.0 & 90.6 \\
         no & 0.169 & 76.0  & 83.9 & 44.5 & 53.5 \\ \bottomrule
    \end{tabular}
    \caption{\textbf{Objective function ablation.} Ablation on choice of objective function. We investigate whether the reversible loss helps with training and generalization. We evaluate on MoCA dataset.}
    \label{tab:moca_ablate}
\end{table}

{
\newpage

\vspace{-0.2cm}
\subsection{Image ordering baselines}
\vspace{-0.1cm}

In the main paper, we compare our method against baselines for ordering image 
sequences. We note that the numbers reported in the paper for differentiable sorting methods use a different backbone compared to ours. We experiment with the same backbone, and the performance remains worse than our method throughout both datasets and all metrics.

\begin{table}[h]
    \centering
    \scriptsize
    \setlength{\tabcolsep}{2pt}
    \begin{tabular}{l|ccc|ccc} \toprule
       & MNIST: 5 & 9 & 16 & SVHN: 5&9&16 \\ \midrule
        DSort & 78.9$|$89.6  & 72.9$|$90.0& 48.9$|$85.5$|$91.1   
          & 34.5$|$63.6 & 8.7$|$53.0 & 0.0$|$36.1$|$40.4 \\ 
        DSv2 & 85.8$|$91.9  & 76.5$|$90.2& 60.7$|$89.0$|$91.6   
          & 43.2$|$69.9 & 16.0$|$60.3 & 0.0$|$45.6$|$54.1 \\
Ours    &    \textbf{93.9}$|$\textbf{96.7}     &    \textbf{87.9}$|$\textbf{95.2}    &    \textbf{72.2}$|$\textbf{91.2}$|$\textbf{92.9}     &     \textbf{77.3}$|$\textbf{88.2}   &  \textbf{53.9}$|$\textbf{81.0}      &  \textbf{19.4}$|$\textbf{67.9}$|$\textbf{67.6}   \\ \bottomrule
    \end{tabular}

    \caption{\textbf{Backbone ablation.} We show that our method remains superior at ordering despite using the same backbone.}
\end{table}

%{ \color{blue}
%\vspace{-0.2cm}
%\subsection{Using pretrained image backbones}
%\vspace{-0.1cm}

%Instead of training the model from scratch, we  also try using DINOv2-S (21M) to extract features, and use the features as input to our model. We empirically found that this is better than initializing the spatial attention layers in our encoder with DINO weights, both frozen and finetuned.

%The experimental settings follow Table 2 (train on MUDS, test on Mono-MUDS).
%The results show better segmentation, and slightly worse localization than ours, suggesting our smaller model is well-trained.
%We also note that pretrained models come in fixed sizes (smallest being 21M), making it much larger than ours, and larger patch size (14px, ours 7px) limits the resolution of localization map.

%We also find that (i) fine-tuning DINO on small dataset is a hindrance (ii) DINO did not help generalize cross-domain.

%\begin{table}[h]

%    \centering
%    \vspace*{-0.8\baselineskip}
%    \footnotesize
%    \setlength{\tabcolsep}{3pt}
%    \begin{tabular}{lccc|ccc} \toprule
     % \multicolumn{3}{c|}{test dataset: Mono-MUDS} & \multicolumn{3}{c}{in-domain (train: MUDS)}  \midrule
%         model & ft & params & FLOPs & order(EM) & loc(acc) & seg(mIoU)  \\ \midrule
%         DINOv2 & no  & 25M & 23G & 86.7 & 80.0 & \textbf{41.3} \\ 
%         DINOv2 & yes  & 25M & 23G & 85.0 & 75.0 & 37.5  \\ 
%         Ours & --  & 4M & 4G& \textbf{90.0} & \textbf{83.3} & 37.1  \\ 
         %Ours-B & --  & 122M & & \\ 
%        \bottomrule
%    \end{tabular}
   % \vspace*{-0.8\baselineskip}
%\end{table}

%}

\newpage

\section{Qualitative Results}
\label{sec:supp7_qualitative}

\subsection{More qualitative results}

We show more qualitative results in Figure \ref{fig:supp}. The model is able to find different cues and use them for ordering.

\begin{figure}[h]
  \centering
  \includegraphics[width=.8\textwidth]{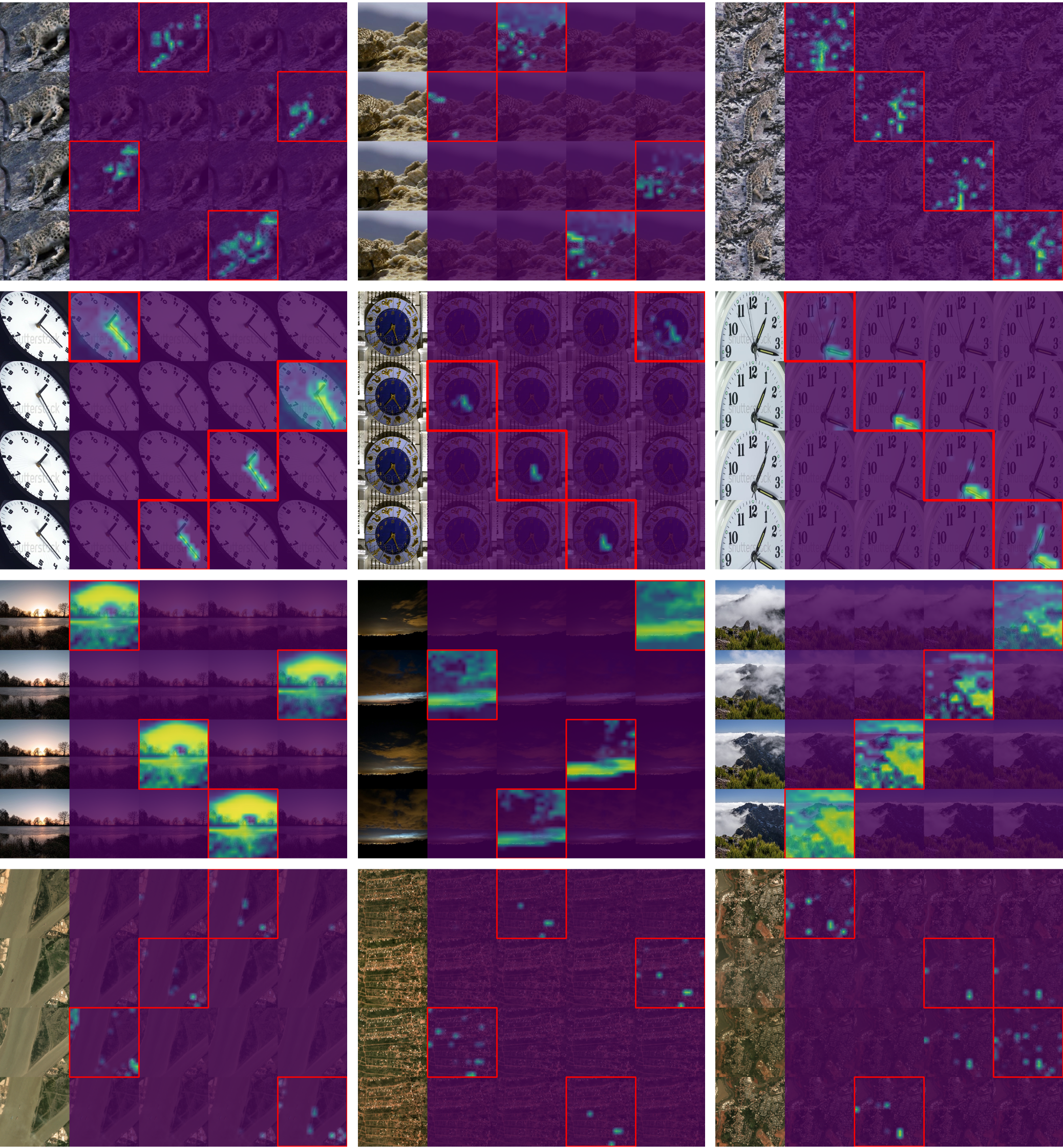}
  \caption{\textbf{Qualitative results.} We show more qualitative results across different datasets. 
  From top to bottom row: Camouflaged animals: the model uses motion as a cue in order to localize the moving subject; Timelapse clocks: the model correctly localizes the clock hands; Timelapse scenes: the model uses either the color of the sky or the objects such as clouds in ordering; Satellite images: the model is able to identify and localize structural changes in landscape such as new buildings and cleared land while being invariant to seasonal changes.
  }
  \label{fig:supp}
\end{figure}
\newpage
\subsection{Unorderable sequences}
\label{sec:supp7_qualitative2}
We show qualitative examples of unorderable sequences in Figure~\ref{fig:supp_fail}. As we simply sample frames from video clips, there will be some sequences that cannot be ordered as they do not contain ordering cues. In particular, this can happen when: (i) there is insufficient change; (ii) changes are stochastic; or (iii) changes are seasonal (cyclic).

  %by looking at the (lack of) attention in the similarity map. %\weidi{we can detect it if the prediction is wrong and by looking at the attention ?}

\begin{figure}[h]
  \centering
  \includegraphics[width=\textwidth]{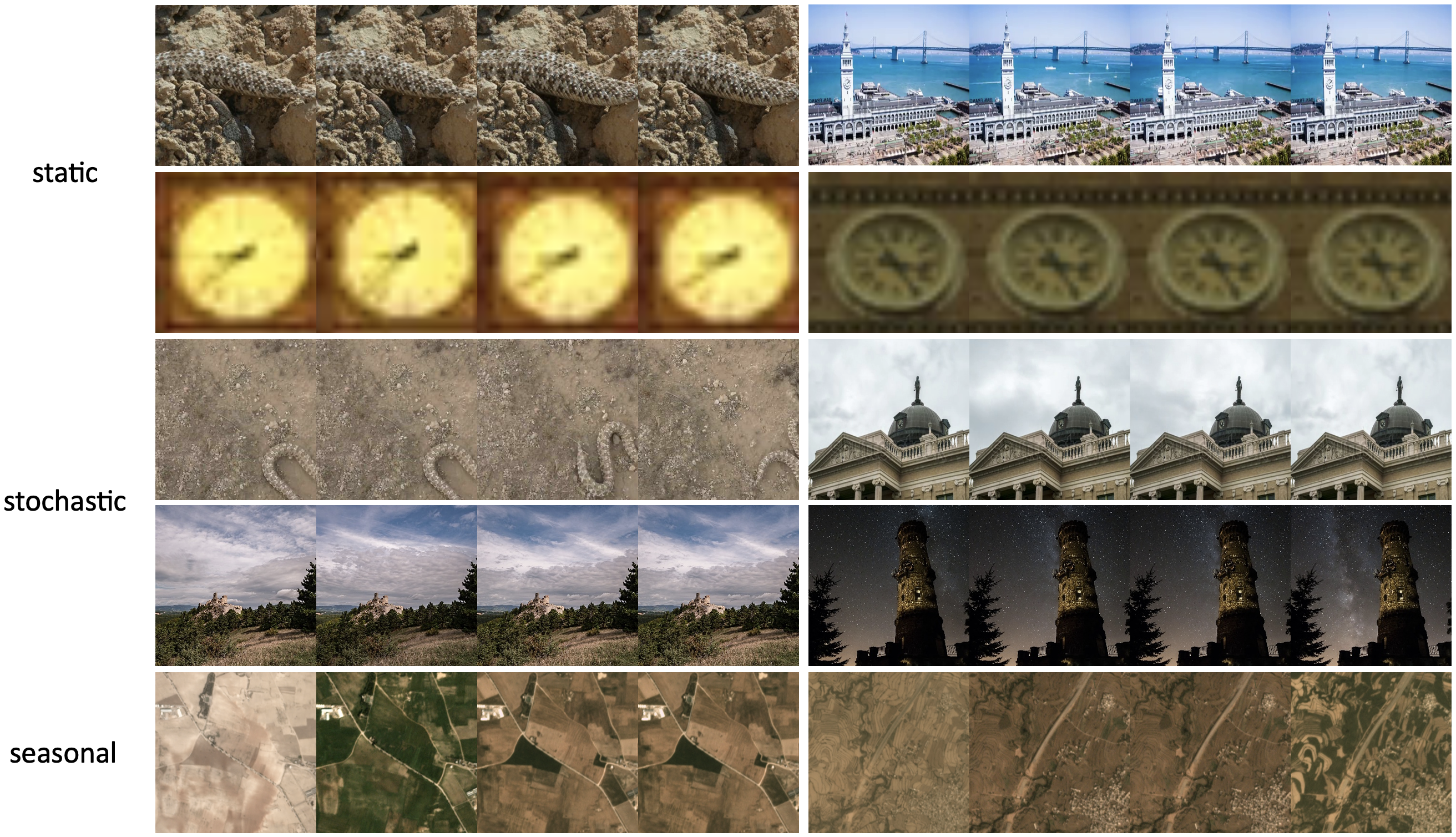}
  \caption{\textbf{Unorderable sequences.} The figure shows examples of unorderable sequences across different datasets.  Those selected are very reasonable: most contain insufficient cues for ordering to be determined, as the cues are either (i) too static, (ii) too stochastic, or (iii) uncorrelated with forward time (e.g. seasonal).
  }
  \label{fig:supp_fail}
\end{figure}

%%%%%%%%%%%%%%%%%%%%%%%%%%%%%%%%%%%%%%%%%%%%%%%%%%%%%%%%%%%%
%{\small
%\bibliographystyle{nips} 
%\bibliography{bib}
%}

%% file: suppfig/pseudocode.tex
  {\ \footnotesize
\begin{verbatim}
class Made2Order(nn.Module):
  def __init__(self):
    self.query = nn.Parameter([1,q,d]).repeat(b,1,1) # (q=queries, b=bsz)
    
  def forward(self, video): 
    # input video of dimensions: b f c h w (f=frames, chw=image dimensions)
    x = patch_posemb(video) # b (f n) d (n=number of tokens)
    x_enc = self.TransformerEncoder(x) # b (f n) d
    x_dec = self.TransformerDecoder(self.query, x_enc) # b q d
    x_enc = F.normalize(x_enc, 2)
    x_dec = F.normalize(x_dec, 2)
    S = torch.einsum(`bik,bjk->bij', x_enc, x_dec) # b (f n) q
    return einops.reduce(S, `b (f n) q -> b f q', `max')      \end{verbatim}
}